\documentclass[sigconf]{acmart}

\usepackage{booktabs} % For formal tables
\usepackage[ruled,lined]{algorithm2e}
\usepackage{natbib}
\usepackage{mathtools}
\usepackage{graphicx}
\usepackage{subfigure}
\usepackage{algorithmic}
\usepackage{censor}

\setcopyright{rightsretained}
\copyrightyear{2020}
\acmYear{2020}
\setcopyright{acmlicensed}\acmConference[GECCO '20]{Genetic and Evolutionary Computation Conference}{July 8--12, 2020}{Cancún, Mexico}
\acmBooktitle{Genetic and Evolutionary Computation Conference (GECCO '20), July 8--12, 2020, Cancún, Mexico}
\acmPrice{15.00}
\acmDOI{10.1145/3377930.3390232}
\acmISBN{978-1-4503-7128-5/20/07}

\graphicspath{ {images/} }

\begin{document}

\title{Covariance Matrix Adaptation for the Rapid Illumination of Behavior Space}

\author{Matthew C. Fontaine}
\affiliation{%
  \institution{Viterbi School of Engineering\\University of Southern California}
  \city{Los Angeles} 
  \state{CA} 
}
\email{mfontain@usc.edu}

\author{Julian Togelius}
\affiliation{%
  \institution{Tandon School of Engineering\\New York University}
  \city{New York City} 
  \state{NY} 
}
\email{julian@togelius.com}

\author{Stefanos Nikolaidis}
\affiliation{%
  \institution{Viterbi School of Engineering\\University of Southern California}
  \city{Los Angeles} 
  \state{CA} 
}
\email{nikolaid@usc.edu}

\author{Amy K. Hoover}
\affiliation{%
  \institution{Ying Wu College of Computing\\New Jersey Institute of Technology}
  \city{Newark} 
  \state{NJ} 
}
\email{ahoover@njit.edu}

\begin{abstract} 
We focus on the challenge of finding a diverse collection of quality solutions on complex continuous domains. While quality diversity (QD) algorithms like Novelty Search with Local Competition (NSLC) and MAP-Elites are designed to generate a diverse range of solutions, these algorithms require a large number of evaluations for exploration of continuous spaces. Meanwhile, variants of the Covariance Matrix Adaptation Evolution Strategy \mbox{(CMA-ES)} are among the best-performing derivative-free optimizers in single-objective continuous domains. This paper proposes a new QD algorithm called Covariance Matrix Adaptation MAP-Elites (CMA-ME). Our new algorithm combines the self-adaptation techniques of CMA-ES with archiving and mapping techniques for maintaining diversity in QD. Results from experiments based on standard continuous optimization benchmarks show that CMA-ME finds better-quality solutions than MAP-Elites; similarly, results on the strategic game Hearthstone show that CMA-ME finds both a higher overall quality and broader diversity of strategies than both CMA-ES and MAP-Elites. Overall, CMA-ME more than doubles the performance of MAP-Elites using standard QD performance metrics. These results suggest that QD algorithms augmented by operators from state-of-the-art optimization algorithms can yield high-performing methods for simultaneously exploring and optimizing continuous search spaces, with significant applications to design, testing, and reinforcement learning among other domains.

\end{abstract}

  % put your semicolon-separated keywords here!

\maketitle

\section{Introduction}

% Replace me
\begin{figure}[t!]
\includegraphics[draft=false,width=0.40\textwidth]{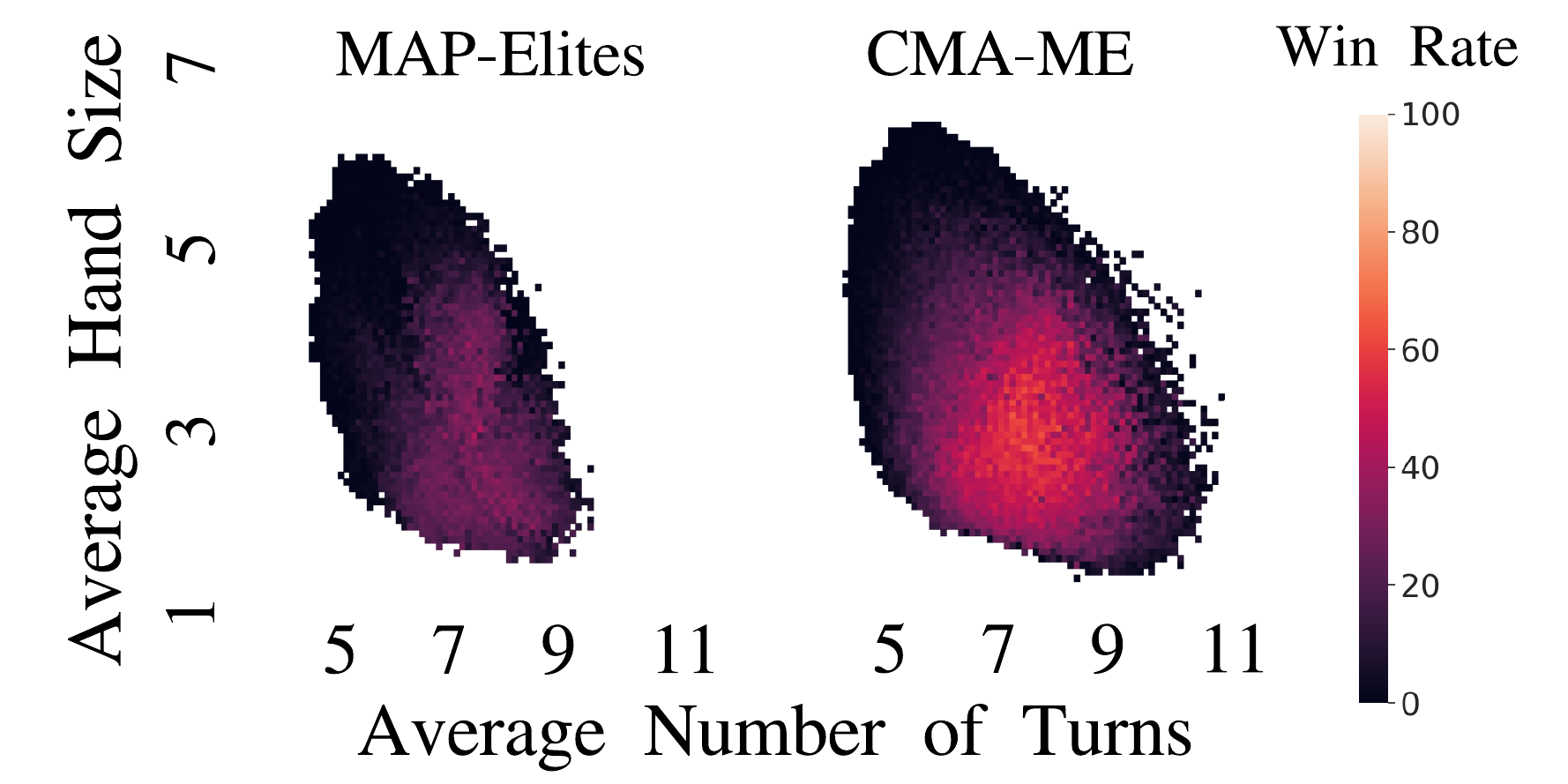}
\caption{\emph{Comparing Hearthstone Archives.} Sample archives for both MAP-Elites and CMA-ME from the Hearthstone experiment. Our new method, CMA-ME, both fills more cells in behavior space and finds higher quality policies to play Hearthstone than MAP-Elites. Each grid cell is an elite (high performing policy) and the intensity value represent the win rate across 200 games against difficult opponents.}
\label{fig:elite-map}
\end{figure}

We focus on the challenge of finding a diverse collection of quality solutions on complex continuous domains. Consider the example application of generating strategies for a turn-based strategy game (e.g., chess, Go, Hearthstone). What makes these games appealing to human players is not the presence of an optimal strategy, but the variety of fundamentally different viable strategies. For example, ``aggressive'' strategies aim to end the game early through high-risk high-reward play, while ``controlling'' strategies  delay the end of the game by postponing the targeting of the game objectives~\cite{de2019evolving}. These example strategies vary in one aspect of their \emph{behavior}, measured by the number of turns the game lasts.

Finding fundamentally different strategies requires navigating a continuous domain, such as the parameter space (weights) of a neural network that maps states to actions, while simultaneously exploring a diverse range of behaviors. Quality Diversity algorithms, such as Novelty Search with Local Competition (NSLC)~\citep{lehman:gecco11} and Multi-dimensional Archive of Phenotypic Elites (MAP-Elites)~\citep{cully:nature15}, drive a divergent search for multiple good solutions, rather than a convergent  search towards a single optimum, through sophisticated archiving and mapping techniques. Solutions are binned in an archive based on their behavior and compete only with others exhibiting similar behaviors. Such stratified competition results in the discovery of potentially sub-optimal solutions called \emph{stepping stones}, which have been shown in some domains to be critical for escaping local optima~\citep{lehman:ec11,gaier:gecco19}. However, these algorithms typically require a large number of evaluations to achieve good results. 

Meanwhile, when seeking a single global optimum, variants of the Covariance Matrix Adaptation Evolution Strategy (CMA-ES)~\cite{hansen:ec01, hansen:cma16} are among the best-performing derivative-free optimizers, with the capability to rapidly navigate continuous spaces. However, the self-adaptation and cumulation techniques driving CMA-ES have yet to successfully power a QD algorithm.

We propose a new hybrid algorithm called Covariance Matrix Adaptation MAP-Elites (CMA-ME), which rapidly navigates and optimizes a continuous space with CMA-ES seeded by solutions stored in the MAP-Elites archive. The hybrid algorithm employs CMA-ES's ability to efficiently navigate continuous search spaces by maintaining a mixture of normal distributions (candidate solutions) dynamically augmented by objective function feedback. The key insight of CMA-ME is to \textit{leverage the selection and adaptation rules of CMA-ES to optimize good solutions, while also efficiently exploring new areas of the search space}.

Building on the underlying structure of MAP-Elites, CMA-ME maintains a population of modified CMA-ES instances called \emph{emitters}, which like CMA-ES operate based on a sampling mean, covariance matrix, and adaptation rules that specify how the covariance matrix and mean are updated for each underlying normal distribution. The population of emitters can therefore be thought of as a Gaussian mixture where each normal distribution focuses on improving a different area of behavior space. This paper explores three types of candidate emitters with different selection and adaptation rules for balancing quality and diversity, called the \emph{random direction}, \emph{improvement}, and \emph{optimizing} emitters. Solutions generated by the emitters are saved in a single unified archive based on their corresponding behaviors.

We evaluate CMA-ME through two experiments: a toy domain designed to highlight current limitations of QD in continuous spaces and a practical turn-based strategy game domain, Hearthstone, which mirrors a common application of QD: finding diverse agent policies.\footnote{Code is available for both the continuous optimization benchmark and Hearthstone domains~\cite{qd-benchmark:github19, evostone:github19}} Hearthstone is an unsolved, partially observable game that poses significant challenges to current AI methods~\cite{hoover:AI19}. Overall, the results of both experiments suggest CMA-ME is a competitive alternative to MAP-Elites for exploring continuous domains (Fig.~\ref{fig:elite-map}). The potential for improving 
QD's growing number of applications is significant as our approach greatly reduces the computation time required to generate a diverse collection of high-quality solutions.

\section{Background}
This section outlines previous advancements in quality diversity (QD) including one of the first QD algorithms, MAP-Elites, and background in CMA-ES to provide context for the CMA-ME algorithm proposed in this paper. 

\subsection{Quality Diversity (QD)}

QD algorithms are often applied in domains where a diversity of good but meaningfully different solutions is valued. For example QD algorithms can build large repertoires of robot behaviors \citep{cully:gecco13,cully:ec16,cully:gecco19} or a diversity of locomotive gaits to help robots quickly respond to damage~\citep{cully:nature15}. By interacting with AI agents, QD can also produce a diversity of generated video game levels~\citep{khalifa:gecco18, alvarez:cog19, gravina:cog19}.

While traditional evolutionary algorithms speciate based on encoding and fitness, a key feature of the precursor to QD (e.g.,\ Novelty Search (NS)~\citep{lehman:alife08}) is speciation through behavioral diversity~\citep{lehman:alife08}. Rather than optimizing for performance relative to a fitness function, searching directly for behavioral diversity promotes the discovery of sub-optimal solutions relative to the objective function. Called \emph{stepping stones}, these solutions mitigate premature convergence to local optima. To promote intra-niche competition \citep{pugh:frontiers16} objectives were reintroduced in the QD algorithms Novelty Search with Local Competition (NSLC) \citep{lehman:gecco11} and MAP-Elites \citep{cully:nature15}.

While NSLC and MAP-Elites share many key features necessary for maintaining a diversity of quality solutions, a core difference between the two is whether the archive is dynamically or statically generated. NSLC dynamically creates behavior niches by growing an archive of sufficiently novel solutions while MAP-Elites (detailed in the next section) maintains a static mapping of behavior. For CMA-ME we choose MAP-Elites as our diversity mechanism to directly compare benefits inherited from CMA-ES. Though we make this design choice for the CMA-ME algorithm solely for comparability, the same principles can be applied to the NSLC archive.

\subsection{MAP-Elites}

While one core difference between two early QD algorithms NSLC \citep{lehman:gecco11} and MAP-Elites~\citep{cully:nature15} is whether behaviors are dynamically or statically mapped, another is the number of behavioral dimensions among which solutions typically vary. Rather than defining a single distance measure to characterize and differentiate behaviors, \mbox{MAP-Elites} often searches along at least two measures called behavior characteristics (BCs) that induce a Cartesian space (called a \emph{behavior space}). This behavior space is then tessellated into uniformly spaced grid cells, where the goal of the algorithm is to 1) maximize the number of grid cells containing solutions and 2) maximize the quality of the best solution within each grid cell. Modifications and improvements to MAP-Elites often focus on the tessellation of behavior space~\citep{vassiliades:ec18,smith:ppsn16,fontaine:gecco19}. 

However, this paper proposes improvements to \mbox{MAP-Elites} based on the generation of solutions rather than the tessellation of behavior space. At the start of the MAP-Elites algorithm, the archive (map) is initialized randomly by solutions sampled uniformly from the search space. 
Each cell of the map contains at most one solution (i.e., an \emph{elite}), which is the highest performing solution in that behavioral niche.
New solutions are generated by taking an elite (selected uniformly at random) and perturbing it with Gaussian noise. MAP-Elites computes a behavior vector for each new solution and assigns the new solution to a cell in the map. The solution replaces the elite in its respective cell if the new solution has higher fitness, or the new solution simply fills the cell if the cell is empty.

\subsection{CMA-ES} \label{subsec:CMA-ES}

Evolution strategies (ES) are a family of evolutionary algorithms that specialize in optimizing continuous spaces by sampling a population of solutions, called a generation, and gradually moving the population toward areas of highest fitness. One canonical type of ES is the  $(\mu / \mu, \lambda)$-ES, where a population of $\lambda$ sample solutions is generated, then the fittest $\mu$ solutions are selected to generate new samples in the next generation. The $(\mu / \mu, \lambda)$-ES recombines the $\mu$ best samples through a weighted average into one mean that represents the center of the population distribution of the next generation. The Covariance Matrix Adaptation Evolution Strategy (CMA-ES) is a particular type of this canonical ES, which is one of the most competitive derivative-free optimizers for single-objective optimization of continuous spaces~\citep{hansen:gecco10}.

CMA-ES models the sampling distribution of the population as a multivariate normal distribution $\mathcal{N}(m, C)$ where $m$ is the distribution mean and $C$ is its covariance matrix. The main mechanisms steering CMA-ES are the selection and ranking of the $\mu$ fittest solutions, which update the next generation's next sampling distribution, $\mathcal{N}(m, C)$. CMA-ES maintains a history of aggregate changes to $m$ called an evolution path, which provides benefits to search that are similar to momentum in stochastic gradient descent.

\subsection{Related Work}

It is important to note that QD methods differ both from diversity maintenance methods and from multi-objective optimization algorithms, in that QD methods search for solutions that exhibit different \textit{behaviors}, which in our strategy game example would be number of turns, rather than searching for diversity in parameter space (neural network weights that induce a strategy). For example, consider the case of two different sets of network weights exhibiting similar play styles. A diversity maintenance method, such as niching or speciation would consider them different species, while QD would treat them as similar with respect to the exhibited behavior, forcing intra-niche competition.  Several versions of CMA-ES exist that incorporate niching or speciation~\citep{shir:icpps06,preuss:ecec12}.

Multi-objective search could also be applied to our strategy game. By treating a behavior characteristic (game length) as an additional objective, we aim to maximize or minimize the average game length in addition to maximizing win rate. However, without any insight into our strategy game, it is unclear whether we should maximize or minimize game length. Quality diversity algorithms differ from multi-objective search by seeking solutions across the whole spectrum of this measure, rather than only at the extremes. 

Several previous works explored incorporating ideas from a simplified ES. For example \citet{conti:nips18} introduced novelty seeking to a \mbox{$(\mu / \mu, \lambda)$-ES}. However, their ES does not leverage adaptation and perturbs solutions through static multivariate Gaussian noise. \citet{nordmoen:alife18} dynamically mutate solutions in \mbox{MAP-Elites}, but globally adapt $\sigma$ (mutation power) for all search space variables rather than  adapting the covariances between search space variables. \citet{vassiliades:gecco18} exploited correlations between elites in \mbox{MAP-Elites}, proposing a variation operator that accelerates the MAP-Elites algorithm. Their approach exploits covariances between elites rather than drawing insights from CMA-ES to model successful evolution steps as a covariance matrix.

\section{Approach: The CMA-ME Algorithm}

Through advanced mechanisms like step-size adaptation and evolution paths, CMA-ES can quickly converge to a single optimum. The potential for CMA-ES to refine the best solutions discovered by MAP-Elites is clear. However, the key insight making CMA-ME possible is that by repurposing these mechanisms from CMA-ES we can improve the \emph{exploration} capabilities of MAP-Elites. Notably, CMA-ES can efficiently both \emph{expand} and contract the search distribution to explore larger regions of the search space and stretch the normal distribution within the search space to find hard to reach nooks and crannies within the behavior space. In CMA-ME we create a population of modified CMA-ES instances called \emph{emitters} that perform search with feedback gained from interacting with the archive.

At a high-level, CMA-ME is a scheduling algorithm for the population of emitters. Solutions are generated in 
search space in a round-robin fashion, where each emitter generates the same number of solutions (see Alg.~\ref{alg:cma-me}). The solutions are generated in the same way for all emitters by sampling from the distribution $\mathcal{N}(m, C)$ (see $\mbox{generate\_solution}$ in Alg.~\ref{alg:cma-me}). The procedure $\mbox{return\_solution}$ is specific to each type of emitter used by CMA-ME and is responsible for adapting the sampling distribution and maintaining the sampled population.

\begin{algorithm}[ht]
\SetAlgoLined
\caption{Covariance Matrix Adaptation MAP-Elites}
\SetKwInOut{Input}{input}
\SetKwInOut{Result}{result}
\SetKwProg{CMAME}{CMA-ME}{}{}
\DontPrintSemicolon
\CMAME{$(evaluate, n)$}
{
\Input{An evaluation function $evaluate$ which computes a behavior characterization and fitness, and a desired number of solutions $n$.}
\Result{Generate $n$ solutions storing elites in a map $M$.}

\BlankLine
Initialize population of emitters E\;

\For{$i\leftarrow 1$ \KwTo $n$}{

Select emitter $e$ from $E$ which has generated the least solutions out of all emitters in $E$\;

\BlankLine

$x_i \gets \mbox{generate\_solution}(e)$\;
$\beta_i, fitness \gets evaluate(x_i)$\;
$\mbox{return\_solution}(e, x_i, \beta_i, fitness)$
}
}
\label{alg:cma-me}
\end{algorithm}

\subsection{CMA-ME Emitters}

To understand how emitters differ from CMA-ES instances, consider that the covariance matrix in CMA-ES models a distribution of possible search directions within the search space. The distribution captures the most likely direction of the next evolution step where fitness increase will be observed. Unlike estimation of multivariate normal algorithms (EMNAs) that increase the likelihood of reobserving the previous best individuals (see Figure~3 of \citet{hansen:cma16}), CMA-ES increases the likelihood of successful future evolution steps (steps that increase fitness). Emitters differ from CMA-ES by adjusting the ranking rules that form the covariance matrix update to maximize the likelihood that future steps in a given direction result in archive improvements. However, there are many ways to rank, leading to many possible types of emitters.

We propose three types of emitters: optimizing, random direction, and improvement. Like CMA-ES, described in Section~\ref{subsec:CMA-ES}, each emitter maintains a sampling mean $m$, a covariance matrix $C$, and a parameter set $P$ that contains additional CMA-ES related parameters (e.g., evolution path). However, while CMA-ES restarts its search based on the best current solution, emitters are differentiated by their rules for restarting and adapting the sampling distribution, as well as for selecting and ranking solutions.

We explore \emph{optimizing emitters} to answer the question: are  restarts alone enough to promote good exploration in CMA-ME as they are in multi-modal methods? An optimizing emitter is almost identical to CMA-ES, differing only in that restart means are chosen from the location of an elite rather than the fittest solution discovered so far. The random direction and improvement emitters are described below in more detail.

To intuitively understand the \emph{random direction} emitters, imagine trying to solve the classic QD maze domain in the dark  \cite{lehman:alife08, lehman:ec11}. Starting from an initial position in the behavior space, random direction emitters travel in a given direction until hitting a wall, at which point they restart search and move in a new direction. While good solutions to the maze and other low-dimensional behavior spaces can be found by random walks, the black-box nature of the forward mapping from search space to behavior makes the inverse mapping equally opaque. Random direction emitters are designed to estimate the inverse mapping of this correspondence problem.

When a random direction emitter restarts, it emulates a step in a random walk by selecting a random direction or bias vector $v_{\beta}$ to move toward in behavior space. To build a covariance matrix such that it biases in direction $v_{\beta}$ at each generation, solutions in search space are mapped to behavior space ($\beta_{i}$). The mean ($m_{\beta}$) of all $\lambda$ solutions is calculated in behavior space and each direction with respect to the mean calculated. Only solutions that improve the archive are then ranked by their projection value against the line $m_{\beta} + v_{\beta}t$. If none of these solutions improve the archive, the emitter restarts from a randomly chosen elite with a new bias vector $v_{\beta}$.

% Interestingly because random emitters choose the bias vector or direction in behavior space to move toward at the beginning of a restart, it necessarily ignores areas of high fitness that it finds along the way. 

%Amy - OLD: Random direction emitters notably cannot change direction between restarts. If new areas of the search space become profitable before a restart, the emitter will ignore the improvement to continue along its bias direction. \emph{Improvement emitters} address this limitation by instead ranking solutions based on the amount each cell in the archive improves over their previous occupant. When determining the amount of improvement, solutions filling empty cells are prioritized over those replacing existing solutions in their niche by ranking them higher in the covariance matrix update. Rather than exploring a fixed direction for an extended period of time, the advantage of improvement emitters is that they can fluidly adjust their goals based on where progress is currently being made. 

 Interestingly because random emitters choose the bias vector or direction in behavior space to move toward at the beginning of a restart, it necessarily ignores areas of high fitness that it finds along the way. Instead, while exploring \emph{improvement emitters} exploit the areas of high fitness by ranking solutions based on the improvement or change in fitness within each niche. When determining the amount of improvement, solutions filling empty cells are prioritized over those replacing existing solutions in their niche by ranking them higher in the covariance matrix update. Rather than exploring a fixed direction for the duration of the run, the advantage of improvement emitters is that they fluidly adjust their goals based on where progress is currently being made. 

Algorithm~\ref{alg:imp} shows the implementation of return\_solution from algorithm~\ref{alg:cma-me} for an improvement emitter. Each solution $x_i$ that has been generated by the emitter maps to a behavior $\beta_i$ and a cell $M[\beta_i]$ in the map. If the cell is empty (line 2), or if $x_i$ has higher fitness than the existing solution in the cell (line 6), $x_i$ is added to the new generation's $parents$ and the map is updated. The process repeats until the generation of $x_i$s reaches size $\lambda$ (line 9), where we adapt the emitter: If we have found parents that improved the map, we rank them (line 11) by concatenating two groups: first the group of parents that discovered new cells in the map, sorted by their fitness, and second the group of parents that improved existing cells, sorted by the increase in fitness over the previous solution that occupied that cell. If we have not found any solutions that improve the map, we restart the emitter (line 15).

\begin{algorithm}[ht]
\caption{An improvement emitter's return\_solution.}
    \label{alg:imp}
\SetKwInOut{Input}{input}
\SetKwInOut{Result}{result}
\SetKwProg{ReturnSolution}{return\_solution}{}{}
\DontPrintSemicolon
\ReturnSolution{$(e, x_i, \beta_i, fitness)$}{
\Input{An improvement emitter $e$, evaluated solution $x_i$, behavior vector $\beta_i$, and fitness.}
\Result{The shared archive $M$ is updated by solution $x_i$. If $\lambda$ individuals have been generated since the last adaptation, adapt the sampling distribution $\mathcal{N}(m, C)$ of $e$ towards the behavioral regions of largest improvement.}
\BlankLine
\nl Unpack the  parents, sampling mean $m$, covariance matrix $C$, and parameter set $P$ from $e$.
\BlankLine
%\nl \If{$x_i$ improves $M[\beta_i]$}{
%\nl $\Delta_i \gets 0$ \\
\nl    \uIf{$M[\beta_i]$ is empty}{
\nl        $\Delta_i \gets fitness$\;
\nl        Flag that $x_i$ discovered a new cell\;
\nl        Add $x_i$ to $parents$\;
        }
 \nl       \ElseIf{$x_i$ improves $M[\beta_i]$}{
 \nl       $\Delta_i \gets fitness - M[\beta_i].fitness$\;
 \nl       Add $x_i$ to $parents$\;
        }

\BlankLine

 \nl \If{$\mbox{sampled population is size } \lambda$}{
 \nl   \uIf{$parents \neq \emptyset$}{
  \nl       Sort $parents$ by (newCell, $\Delta_i$) \;
    
 \nl       Update $m$, $C$, and $P$ by $parents$ \;
 \nl       $parents \gets \emptyset$\;
    }
 \nl  \Else{
 \nl       Restart from random elite in $M$ \;

    }
}
}
\end{algorithm}

\section{Toy Domain} \label{sec:toy_domain}

%OLD - Previous work on applying QD to Hearthstone deck search~\citep{fontaine:gecco19} observed highly distorted behavior spaces. However, to our knowledge, no previous work involving standard benchmark domains observes distortions in behavior space. The goal of the toy domain is to create the simplest domain with high degrees of behavior space distortions. Surprisingly, a linear projection from a high-dimensional search space to a low-dimensional behavior space highly distorts the distribution of solutions in behavior space. This section details experiments in our toy domain designed to measure the effects of distortions on MAP-Elites. We hypothesize the most likely benefit of CMA-ME is the ability to overcome distortions in behavior space and this experiment measures the benefits of an adaptable QD algorithm over using a fixed mutation power.

Previous work on applying QD to Hearthstone deck search~\citep{fontaine:gecco19} observed highly distorted behavior spaces. However, to our knowledge, no previous work involving standard benchmark domains observes distortions in behavior space. The goal of the toy domain is to create the simplest domain with high degrees of behavior space distortions. Surprisingly, a linear projection from a high-dimensional search space to a low-dimensional behavior space highly distorts the distribution of solutions in behavior space. This section details experiments in our toy domain designed to measure the effects of distortions on MAP-Elites. We hypothesize the most likely benefit of CMA-ME is the ability to overcome distortions in behavior space and this experiment measures the benefits of an adaptable QD algorithm over using a fixed mutation power.

\subsection{Distorted Behavior Spaces} \label{subsec:bates}

Previous work in QD measures the effects of different types of behavior characteristics (BCs) on QD performance, such as BC alignment with the objective~\citep{pugh:gecco15,pugh:frontiers16,fioravanzo:arxiv19}. We highlight a different limiting effect on performance: the distortion caused by dimensionality reduction from search space to behavior space. While any dimensionality reduction can increase the difficulty of covering behavior space, we demonstrate that exploring the behavior space formed from a simple linear projection from a high-dimensional search space results in significant difficulty in standard MAP-Elites.

\begin{figure}
\includegraphics[draft=false,width=0.30\textwidth]{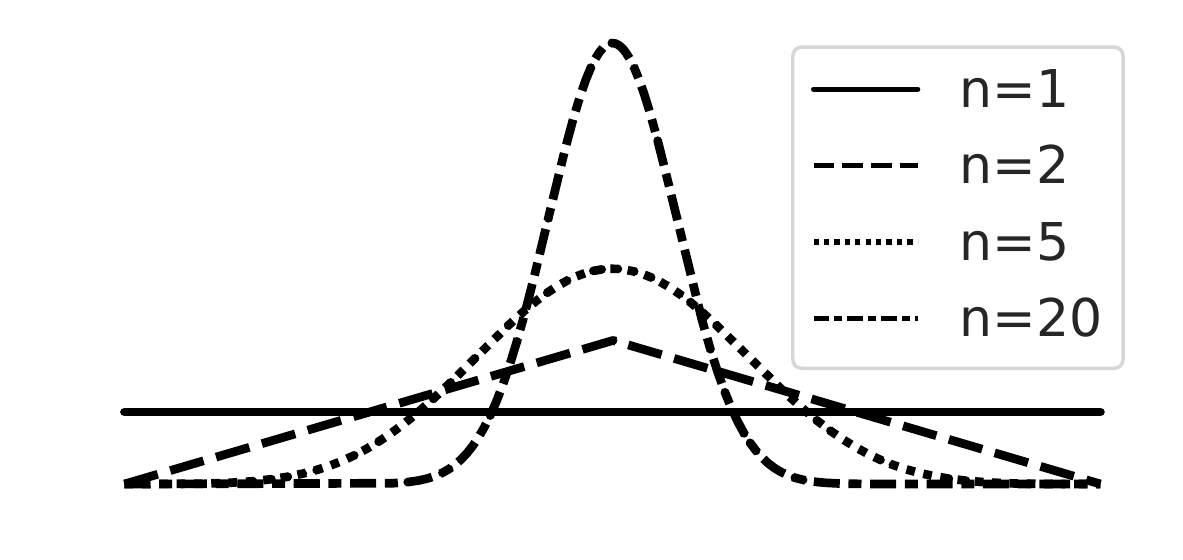}
\caption{A Bates distribution demonstrating the narrowing property of behavior spaces formed by a linear projection.}
\label{fig:bates-distribution}
\end{figure}

Specifically in the case of linear projection, each BC depends on every parameter of the search space. In the case when all projection coefficients are equal, each parameter contributes equally to the corresponding BC. By being equally dependent on each parameter, a QD algorithm needs to navigate every parameter to reach extremes in the behavior space instead of only a subset of the parameters.

This is shown by uniformly sampling from the search space and projecting the samples to behavior vectors, where each component is the sum of $n$ uniform random variables. When divided by $n$ (to normalize the BC to the range $[0,1]$), the sampling results in the Bates distribution shown in Fig.~\ref{fig:bates-distribution}. As the dimensions of the search space grow, the distribution of the behavior space narrows making it harder to find behaviors in the tails of the distribution.

We hypothesize that the adaptation mechanisms of CMA-ME will better cover this behavior space when compared to MAP-Elites, since CMA-ME can adapt each parameter with a separate variance, rather than with a fixed global mutation rate. Additionally, the final goal of this experiment is to explore the performance of these algorithms in a distributed setting, therefore we choose parameters that allow for parallelization of the evaluation.

\subsection{Experiments}

\begin{table*}[]
\caption{Sphere Function Results}
\label{tab:sphere}
\begin{tabular}{l|ccc|ccc}
             & \multicolumn{3}{l|}{$n=20$}          & \multicolumn{3}{l}{$n=100$}  \\ 
    \toprule
Algorithm    & Max Fitness & Cells Occupied & QD-Score & Max Fitness & Cells Occupied & QD-Score \\
    \midrule
CMA-ES       & \bf{100} & 3.46 \% & 731,613 & \bf{100} & 3.74 \% & 725,013 \\
MAP-Elites   & 99.596 & 56.22 \% & 11,386,641 & 96.153 & 26.97 \% & 5,578,919 \\
ME (line)    & 99.920 & 68.99 \% & 13,714,405 & 98.021 & 31.75 \% & 6,691,582 \\
CMA-ME~(opt) & \bf{100} & 12.53 \% & 2,573,157 & \bf{100} & 2.70 \% & 654,649 \\
CMA-ME~(rd)  & 98.092 & \bf{90.32} \% & 13,651,537 & 96.731 & \bf{77.12} \% & \bf{13,465,879}\\
CMA-ME~(imp) & 99.932 & 87.75 \% & \bf{16,875,583} & 99.597 & 61.98 \% & 12,542,848 \\     
  \bottomrule
\end{tabular}
\end{table*}

\begin{table*}[]
\caption{Rastrigin Function Results}
\label{tab:rastrigin}
\begin{tabular}{l|ccc|ccc}
             & \multicolumn{3}{l|}{$n=20$}          & \multicolumn{3}{l}{$n=100$}  \\ 
    \toprule
Algorithm    & Max Fitness & Cells Occupied & QD-Score & Max Fitness & Cells Occupied & QD-Score \\
    \midrule
CMA-ES       & \bf{99.982} & 4.17 \% & 818,090 & \bf{99.886} & 3.64 \% & 660,037 \\
MAP-Elites   & 90.673 & 55.70 \% & 9,340,327 & 81.089 & 26.51 \% & 4,388,839 \\
ME (line)    & 92.700 & 58.25 \% & 9,891,199 & 84.855 & 27.72 \% & 4,835,294 \\
CMA-ME~(opt) & 99.559 & 8.63 \% & 1,865,910 & 98.159 & 3.23 \% & 676,999 \\
CMA-ME~(rd)  & 91.084 & \bf{87.74 \%} & 10,229,537 & 90.801 & \bf{74.13 \%} & \bf{10,130,091} \\
CMA-ME~(imp) & 96.358 & 83.42 \% & \bf{14,156,185} & 86.876 & 60.72 \% & 9,804,991 \\     
  \bottomrule
\end{tabular}
\end{table*}

To show the benefit of covariance matrix adaptation in \mbox{CMA-ME}, we compare the performance of MAP-Elites, CMA-ES, and \mbox{CMA-ME} when performing dimensionality reduction from the search space to behavior space. We additionally compare against the recently proposed line-mutation operator for MAP-Elites~\cite{vassiliades:gecco18}, which we call ME (line)~\cite{qd-benchmark:github19}. As QD algorithms require a solution quality measure, we include two functions from the continuous black-box optimization set of benchmarks~\citep{hansen:arxiv16,hansen:gecco10} as objectives.

Our two objective functions are of the form $f: {\rm I\!R} ^ n \to {\rm I\!R}$: a sphere shown in Eq.~\ref{eq:sphere} and the Rastrigin function shown in Eq.~\ref{eq:rastrigin}. The optimal fitness of $0$ in these functions is obtained by $x_i = 0$. To avoid having the optimal value at the center of the search space, we offset the fitness function so that, without loss of generality, the optimal location is $x_i = 5.12 \cdot 0.4 = 2.048$ (note that $[-5.12, 5.12]$ is the typical valid domain of the Rastrigin function).

\begin{equation} \label{eq:sphere}
sphere(x) = \sum_{i=1}^{n} {x_i} ^ {2}
\end{equation}

\begin{equation} \label{eq:rastrigin}
rastrigin(x) = 10n + \sum_{i=1}^{n} {[{x_i}^{2} - 10 \cos(2 \pi x_i)]}
\end{equation}

Behavior characteristics are formed by a linear projection from ${\rm I\!R} ^ n$ to ${\rm I\!R} ^ 2$, and the behavior space is bounded through a \emph{clip} function (Eq.~\ref{eq:clip}) that restricts the contribution of each component $x_i$ to the range $[-5.12, 5.12]$ (the typical domain of the constrained Rastrigin function). To ensure that the behavior space is equally dependent on each component of the search space (i.e., ${\rm I\!R} ^ n$), we assign equal weights to each component. The function $p: {\rm I\!R} ^ n \to {\rm I\!R} ^ 2$ formalizes the projection from the search space ${\rm I\!R} ^ n$ to the behavior space ${\rm I\!R} ^ 2$ (see Eq.~\ref{eq:projection}), by computing the sum of the first half of components from ${\rm I\!R} ^ n$ and the sum of the second half of components from ${\rm I\!R} ^ n$. 

\begin{equation} \label{eq:clip}
  clip(x_i) =
  \begin{cases}
    x_i & \text{if $-5.12 \leq x_i \leq 5.12$} \\
    5.12 / x_i & \text{otherwise}
  \end{cases}
\end{equation}

\begin{equation} \label{eq:projection}
p(x) = \left( \sum_{i=1}^{\lfloor{\frac{n}{2}}\rfloor}  {clip(x_i)}, \sum_{i=\lfloor{\frac{n}{2}}\rfloor+1}^{n}  {clip(x_i)} \right)
\end{equation}

We compare MAP-Elites, ME (line) CMA-ES, and CMA-ME on the toy domain, running CMA-ME three times, once for each emitter type. We run each algorithm for 2.5M evaluations. We set $\lambda = 500$ for CMA-ES, whereas a single run of CMA-ME deploys 15 emitters of the same type with $\lambda = 37$~\footnote{Both the Toy and Hearthstone domains use the same $\lambda$ and emitter count.}. The map resolution is $500 \times 500$.  

Algorithms are compared by the QD-score metric proposed by previous work~\cite{pugh:gecco15}, which in MAP-Elites and CMA-ME is the sum of fitness values of all elites in the map. To compute the QD-score of \mbox{CMA-ES}, solutions are assigned a grid location on what their BC would have been and populate a pseudo-archive.  Since \mbox{QD-score} assumes maximizing test functions with non-negative values, fitness is normalized to the range $[0,100]$, where $100$ is optimal. Since we can analytically compute the boundaries of the space from the linear projection, we also show the percentage of possible cells occupied by each algorithm. 

MAP-Elites perturbs solutions with Gaussian noise scaled by a factor $\sigma$ named mutation power. Previous work  \citep{nordmoen:ec18, eiben:ec99} shows that varying $\sigma$ can greatly affect both the precision and coverage of \mbox{MAP-Elites}. To account for this and obtain the best performance of \mbox{MAP-Elites} on the toy domain, we ran a grid search to measure \mbox{MAP-Elites} performance across 101 values of $\sigma$ uniformly distributed across $[0.1, 1.1]$, and we selected the value with the best coverage, $\sigma = 0.5$, for all the experiments. Since CMA-ES and CMA-ME adapt their sampling distribution, we did not tune any parameters, but we set the initial value of their mutation power also to $\sigma = 0.5$. 

\subsection{Results}

We label CMA-ME~(opt), CMA-ME~(rd), and CMA-ME~(imp) for the optimizing, random direction, and improvement emitters, respectively. Tables~\ref{tab:sphere} and~\ref{tab:rastrigin} show the results of the sphere and rastrigin function experiments. CMA-ES outperforms all other algorithms in obtaining the optimal fitness. As predicted, covering the behavior space becomes harder as the dimensions of the search space grow. CMA-ME~(rd) and CMA-ME~(imp) obtain the highest QD-Scores and fill the largest number of unique cells in the behavior space. This is because of the ability of the CMA-ME emitters to efficiently discover new cells.  Notably, CMA-ME~(opt) fails to keep up with CMA-ES, performing worse in maximum fitness. We find this result consistent with the literature on multi-modal CMA-ES~\cite{hansen2004evaluating}, since CMA-ES moves with a single large population that has global optimization properties, while CMA-ME (opt) has a collection of smaller populations with similar behavior.

%CMA-ME~(rd) finds the largest number of unique cells, though it scores the lowest on the maximum fitness metric for $n=20$. However, when the dimensions grow to $n=100$, \mbox{MAP-Elites} achieves the lowest maximum fitness. As predicted, covering the behavior space becomes harder as the dimensions of the search space grow. CMA-ME~(rd) and CMA-ME~(imp) obtain the highest QD-Scores and fill the largest number of unique cells in the behavior space. Notably, CMA-ME~(opt) fails to keep up with CMA-ES, performing worse in all metrics in all experiments (except maximum fitness on Sphere $n=100$). We find this result consistent with the literature on multi-modal CMA-ES~\cite{hansen2004evaluating}, since CMA-ES moves with a large population that has global optimization properties, while CMA-ME (opt) has a collection of smaller populations with similar behavior.

%Table~\ref{tab:rastrigin} shows the results of the Rastrigin function experiments.
%CMA-ES obtains the highest maximum fitness for all experiments. As with the sphere experiment, CMA-ME~(rd) and CMA-ME~(imp) outperform MAP-Elites in obtaining the highest QD-Scores and discovering the most unique cells in behavior space. 

\begin{figure*}
\centering
\subfigure[Sphere Function $n=100$]
{
\includegraphics[draft=false,width=0.23\textwidth]{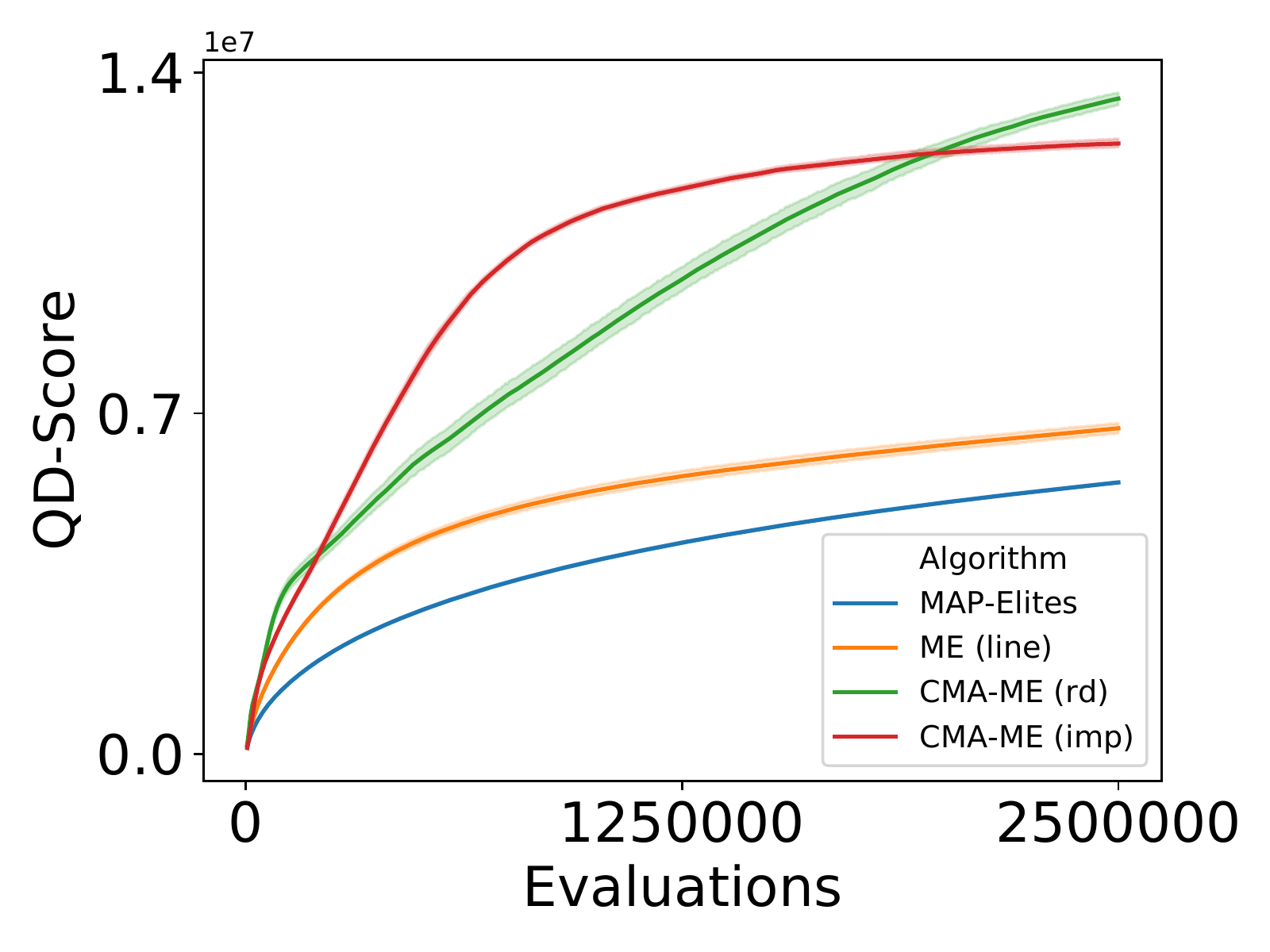}
\label{fig:sphere_qd_score}
}
\subfigure[Rastrigin Function $n=100$]
{
\includegraphics[draft=false,width=0.23\textwidth]{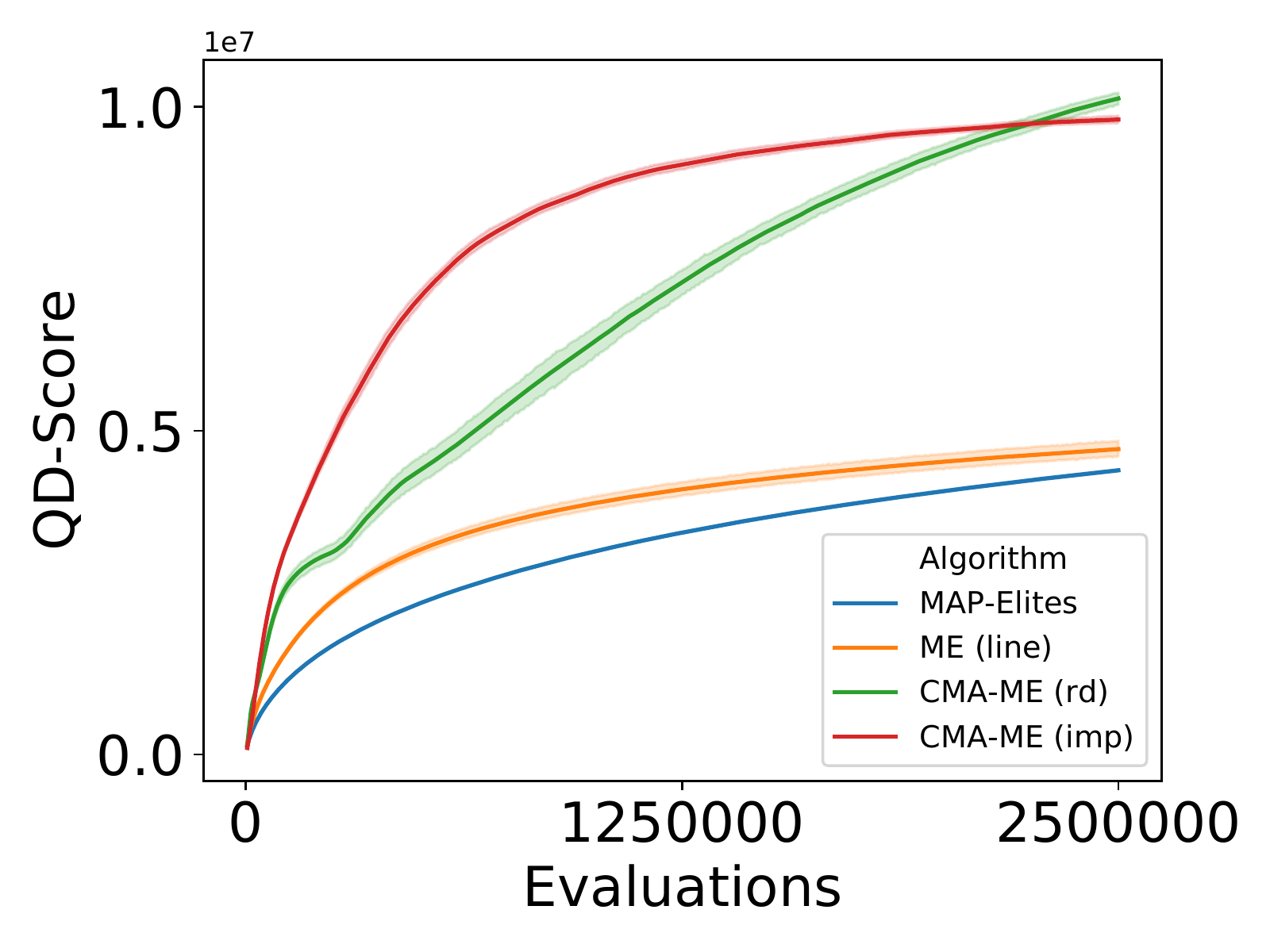}
\label{fig:rastrigin_qd_score}
}
\subfigure[Sphere Function $n=100$ Elites]
{
\includegraphics[draft=false,width=0.23\textwidth]{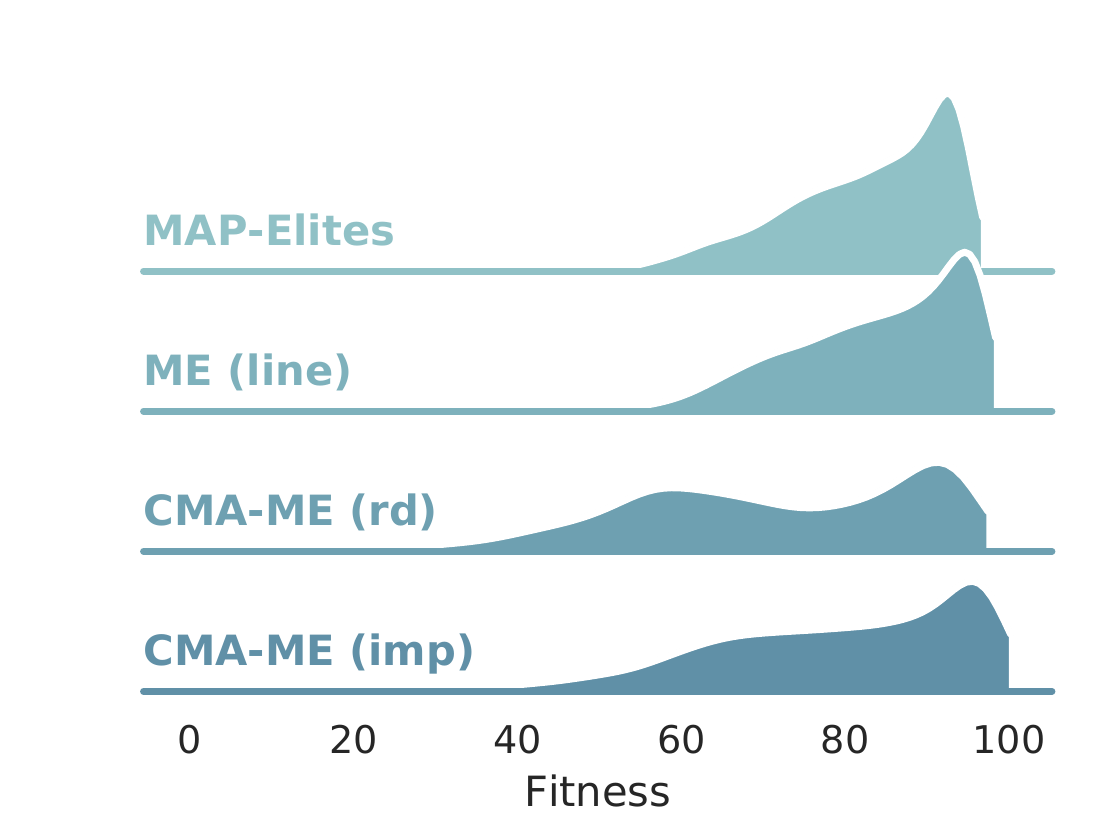}
\label{fig:sphere_elites}
}
\subfigure[Rastrigin Function $n=100$ Elites]
{
\includegraphics[draft=false,width=0.23\textwidth]{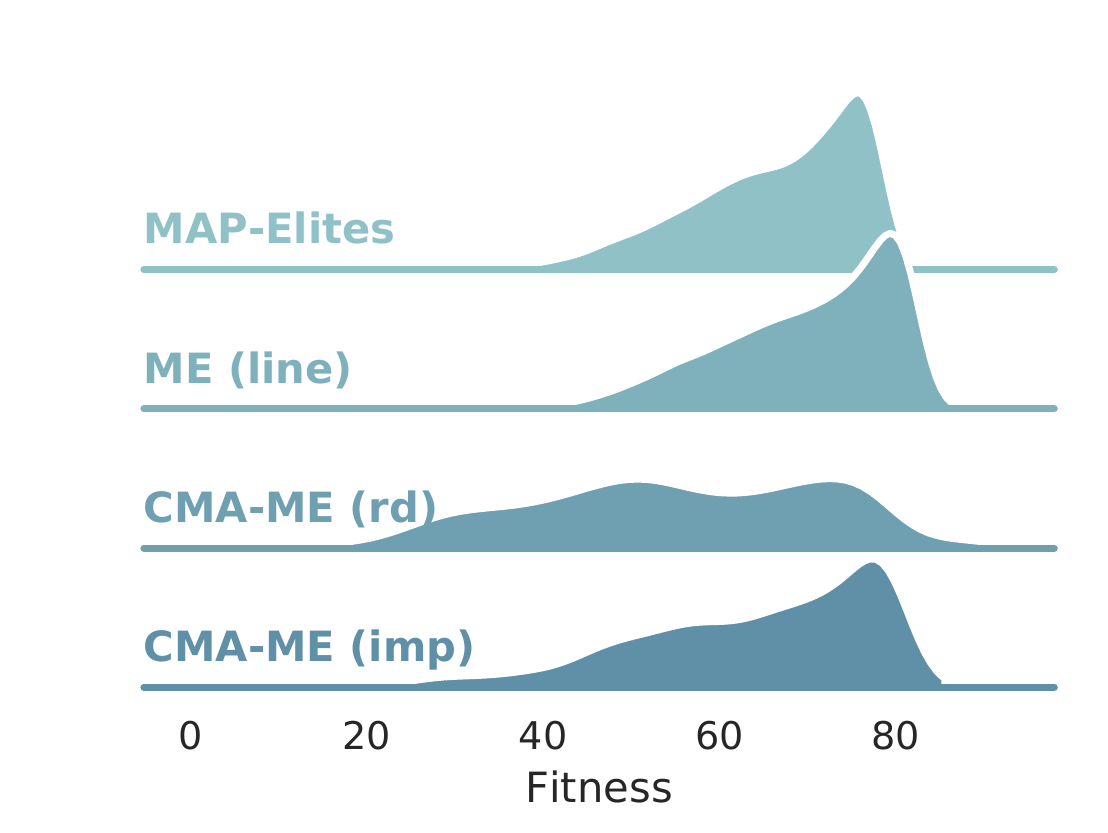}
\label{fig:rastrigin_elites}
}

\caption{\emph{Toy Domain Results.} (a-b) Improvement in QD-Score over evaluations. (c-d) The distribution of elites scaled by the number of occupied cells to show the relative makeup of elites within each archive.
\label{fig:toy-results}}
\end{figure*}

Fig.~\ref{fig:sphere_elites} and Fig.~\ref{fig:rastrigin_elites} show the distribution of elites by their fitness. CMA-ME~(rd) achieves a high QD-score by retaining a large number of relatively low quality solutions, while \mbox{CMA-ME~(imp)} fills slightly less cells but of higher quality. In the Hearthstone experiment (Section~\ref{sec:Hearthstone}) we use improvement emitters, since we are more interested in elites of \emph{high quality} than the number of cells filled in the resulting collection.

%For $n=100$, random direction emitters perform better on both the cells occupied and QD-score metrics. This is because of the ability of the CMA-ME emitters to discover new cells; 

% Overall, these results show that for distorted behavior spaces spaces, CMA-ME~(rd) and CMA-ME~(imp) outperform MAP-Elites, while CMA-ES's best solutions achieve the highest overall fitness. Fig.~\ref{fig:toy-results} shows that CMA-ME~(rd) achieves a high QD-score by retaining a large number of relatively low quality solutions, while \mbox{CMA-ME~(imp)} fills slightly less cells but of higher quality. In the Hearthstone experiment (Section~\ref{sec:Hearthstone}) we use improvement emitters, since we are more interested in elites of \emph{high quality} than the number of cells filled in the resulting collection.

\begin{figure*}[t!]
\centering
\subfigure[Distribution of Elites]
{
\includegraphics[draft=false,width=0.26\textwidth]{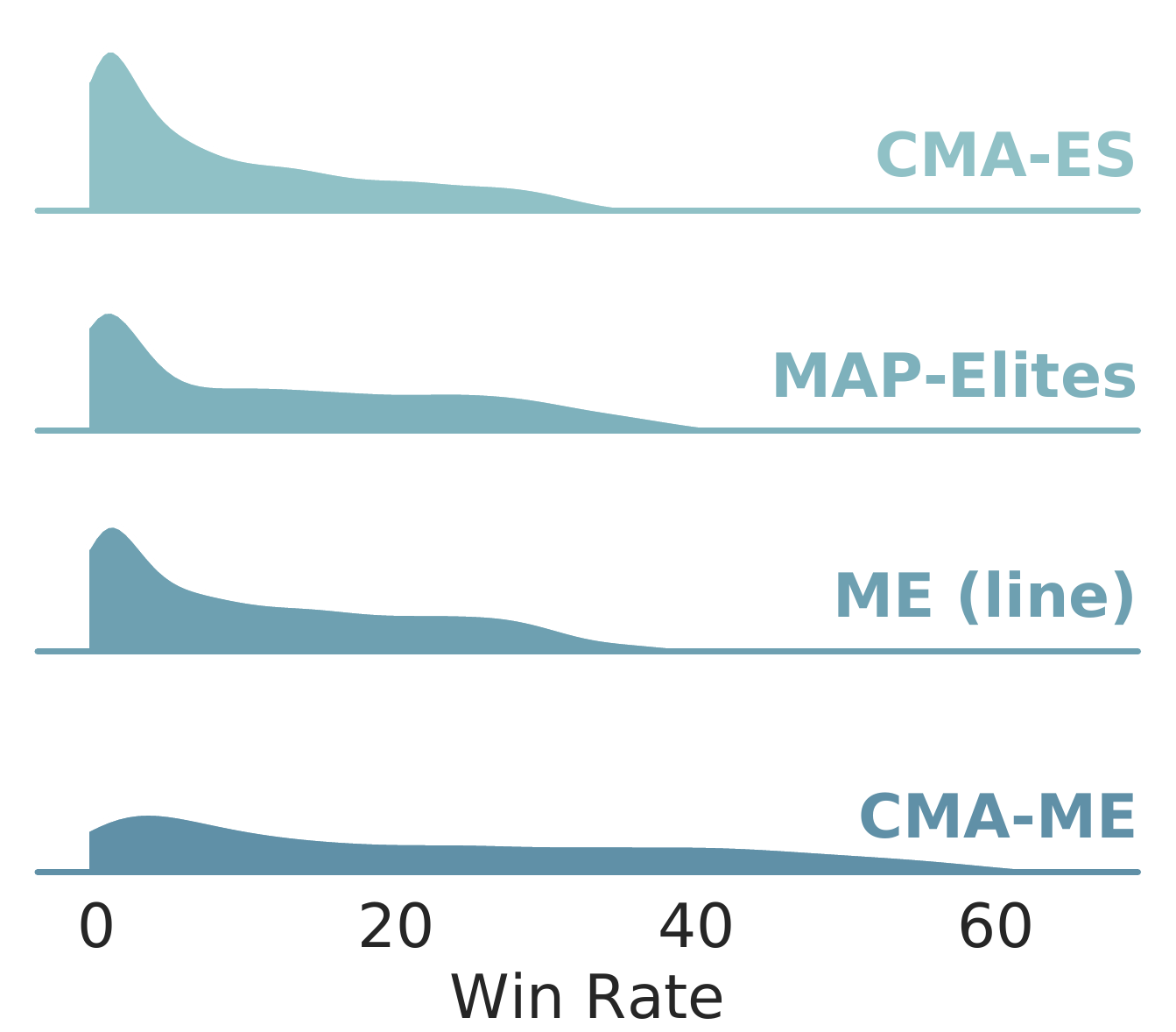}
\label{fig:rogue_elites}
}
\subfigure[QD-Score]
{
\includegraphics[draft=false,width=0.28\textwidth]{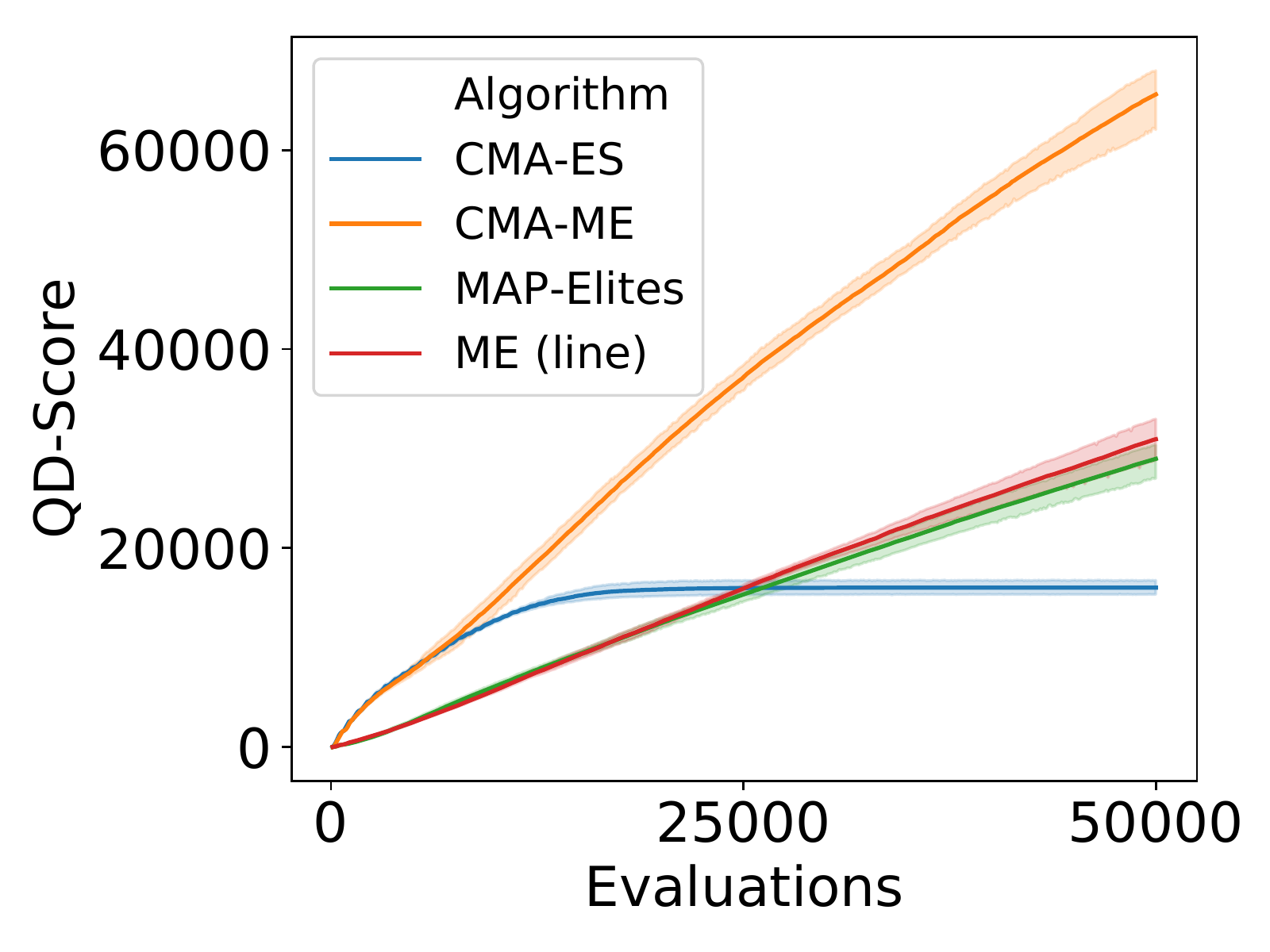}
\label{fig:rogue_qd_score}
}
\subfigure[Win Rate]
{
\includegraphics[draft=false,width=0.28\textwidth]{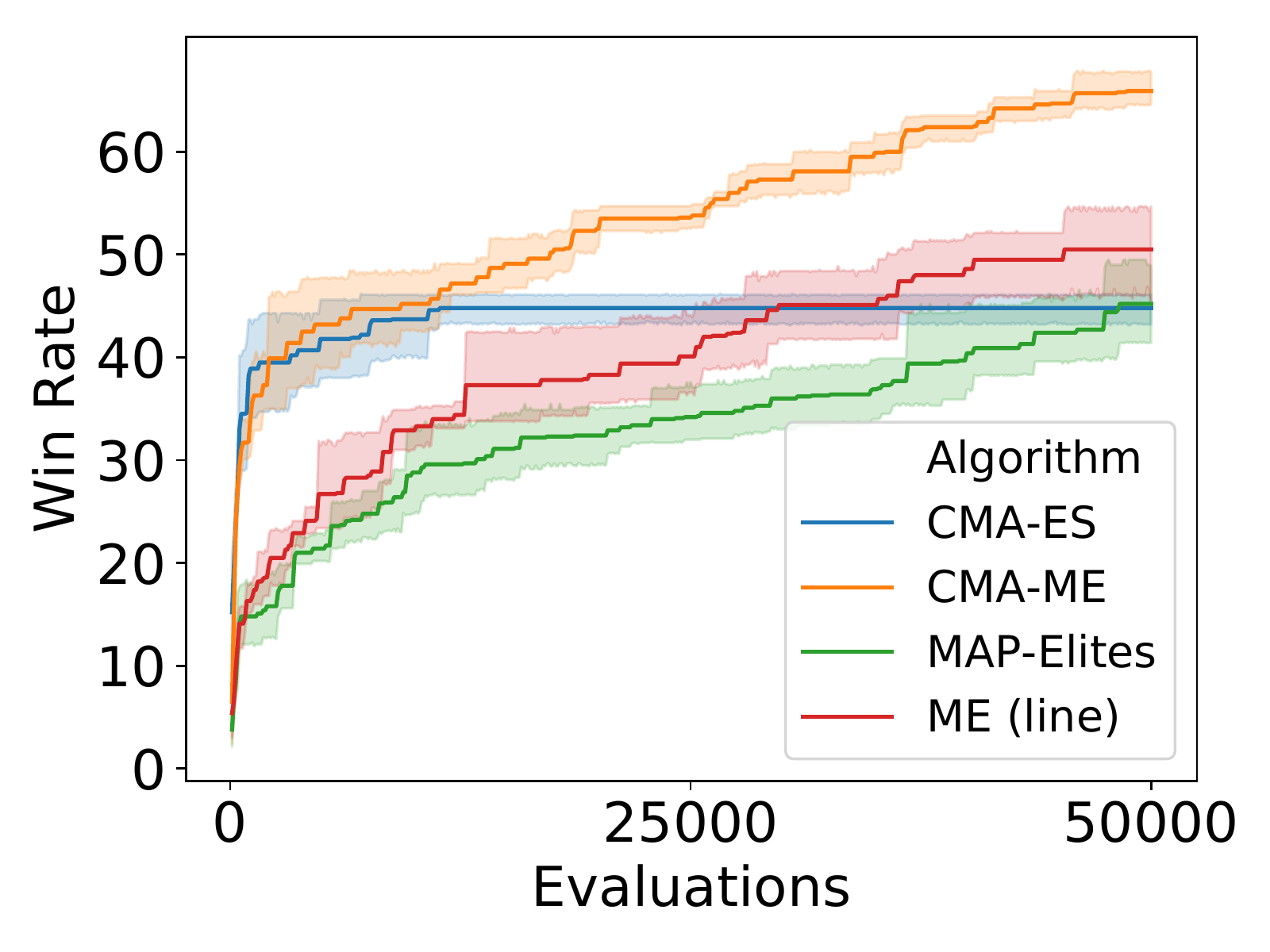}
\label{fig:rogue_win_rate}
}
\caption{\emph{Hearthstone Results} (a) The distribution of elites by win rate. Each distribution is scaled by the number of occupied cells to show the relative makeup of elites within each archive. (b) Improvement in QD-Score over evaluations. (c) Improvement in win rate over evaluations.
\label{fig:hearthstone-results}}
\end{figure*}

\section{Hearthstone Domain} \label{sec:Hearthstone}
We aim to discover a collection of high quality strategies for playing the Hearthstone game against a set of fixed opponents. We wish to explore \emph{how differently} the game can be played rather than just \emph{how well} the game can be played. We selected this domain, since its behavior space has distortions similar to those described in section~\ref{sec:toy_domain} and observed in previous work~\cite{fontaine:gecco19}, which make exploration of the behavior space particularly challenging. 

Hearthstone~\citep{hearthstone:web19} is a two-player, turn-taking adversarial online collectable card game that is an increasingly popular domain for evaluating both classical AI techniques and modern deep reinforcement learning approaches due to the many unique challenges it poses (e.g., large branching factor, partial observability, stochastic actions, and difficulty with planning under uncertainty)~\citep{hoover:AI19}. In Hearthstone players construct a deck of exactly thirty cards that players place on a board shared with their opponent. The game's objective is to reduce your opponent's health to zero by attacking your opponent directly using cards in play. Decks are constrained by one of nine possible hero classes, where each hero class can access different cards and abilities.

Rather than manipulating the reward function of individual agents in a QD system ~\citep{ng:icml99,arulkumaran:geccocomp19} (like the QD approach in AlphaStar \citep{yinyals:nature19}), generating the best gameplay strategy or deck~\citep{garcia:kbs19,swiechowski:cig18,bhatt:fdg18}, or searching for different decks with a fixed strategy \citep{fontaine:gecco19}, our experiments search for a diversity of strategies for playing an expert-level human-constructed deck.

\subsection{Experiments}

This section details the Hearthstone simulator, agent policy and deck, and our opponents' policies and decks.

\noindent\textbf{SabberStone Simulator:} SabberStone~\citep{sabberstone:github19} is a Hearthstone simulator that replicates the rules of Hearthstone and uses the card definitions publicly provided by Blizzard.  In addition to simulating the game, SabberStone includes a turn-local game tree search that searches possible action sequences that can be taken at a given turn. To implement different strategies, users can create a heuristic scoring function that evaluates the state of the game. Included with the simulator are standard ``aggro and control'' card-game strategies which we use in our opponent agents.

\noindent\textbf{Our Deck:} In Hearthstone there are subsets of cards that can only be used by specific ``classes'' of players.  We selected the class Rogue,  where ``cheap'' cards playable in the beginning of the game can be valuable later on. Successful play with the Rogue class requires long-term planning with sparse rewards. To our knowledge, our work is the first to create a policy to play the Rogue class. We selected the Tempo Rogue archetype from the Hearthstone expansion \emph{Rise of Shadows}, which is a hard deck preferred by advanced players. While many variants of the Tempo Rogue arche\-type exist, we decided to use the deck\-list from Hearthstone grandmaster Fei "ETC" Liang who reached the number 1 ranking with his list in May 2019~\citep{roguedeck:web19}.

\noindent\textbf{Opponents:} \citet{fontaine:gecco19} used game tree search to find decks and associated policies. They found six high performing decks for the Paladin, Warlock, and Hunter classes playing aggro and control strategies; we use these as our opponent suite.

\noindent\textbf{Neural Network:} We search for the parameters of a neural network that scores an observable game state based on the cards played on the board and card-specific features. The network maps 15 evaluation features defined by~\citet{sabberstone:github19} to a scalar score. \citet{cuccu:aamas19} show that a six-neuron neural network trained with a natural evolution strategies (NES)~\citep{glasmachers:gecco10}, can obtain competitive and sometimes state-of-the-art solutions on the Atari Learning Environment (ALE)~\citep{bellemare:ai13}. They separate feature extraction from decision-making, using vector quantization for feature extraction and the six-neuron network for the decision making. Motivated by this work, we use a 26 node fully connected feed-forward neural network with layer sizes $[15,5,4,1]$ (109 parameters). 

\subsection{Search Parameters and Tuning:} We use the fitness function proposed in \citep{bhatt:fdg18, fontaine:gecco19}: the average health difference between players at the end of the game, as a smooth approximation of win rate. For MAP-Elites and CMA-ME, we characterize behavior by the average hand size per turn and the average number of turns the game lasts. We choose these behavioral characteristics to capture a spectrum of strategies between aggro decks, which try to end the game quickly, and control decks that attempt to extend the game. The hand size dimension measures the ability of a strategy to generate new cards.

To tune MAP-Elites, we ran three experiments with $\sigma$ values of 0.05, 0.3, and 0.8. MAP-Elites achieved the best coverage and maximum win rate performance with $\sigma = 0.05$ and we used that as our mutation power. Our archive was resolution $100 \times 100$, where we set the range of behavior values using data from the Hearthstone player data corpus~\citep{hsreplay:web19}. As with the Toy Domain in section~\ref{sec:toy_domain}, CMA-ES and CMA-ME used the same hyperparameters as MAP-Elites. For CMA-ME we ran all experiments using improvement emitters, since their performance had the most desirable performance attributes in the toy domain.

\subsection{Distributed Evaluation}

We ran our experiments on a high-performance cluster with 500 (8 core) CPUs in a distributed setting, with a master search node and 499 worker nodes. Each worker node is responsible for evaluating a single policy at a time and plays 200 games against our opponent suite. A single experimental trial evaluating 50,000 policies takes ~12 hours. MAP-Elites and ME (line) are run asynchronously on the master search node, while CMA-ME and CMA-ES synchronize after each generation. We ran each algorithm for 5 trials and generated $50,000$ candidate solutions per trial.

\subsection{Results}

 Table~\ref{tab:hearthstone} shows that CMA-ME outperforms both \mbox{MAP-Elites} and \mbox{CMA-ES} and in maximum fitness, maximum win rate, the number of cells filled, and QD-Score. The distribution of elites for all three algorithms show that CMA-ME finds elites in higher performing parts of the behavior space than both CMA-ES and \mbox{MAP-Elites} (see Fig.~\ref{fig:rogue_elites}). The sample archive for CMA-ME and \mbox{MAP-Elites} in Fig.~\ref{fig:elite-map} similarly illustrates that CMA-ME better covers the behavior space and finds higher quality policies than \mbox{MAP-Elites} and ME (line).
 
 Fig.~\ref{fig:rogue_qd_score} shows the increase in quality diversity over time, with CMA-ME more than doubling the QD-Score of MAP-Elites. Fig.~\ref{fig:rogue_win_rate} shows the increase in win rate over time. CMA-ME maintains a higher win rate than CMA-ES, MAP-Elites and ME (line) at all stages of evaluation. CMA-ES quickly converges to a single solution but is surpassed by MAP-Elites later in the evaluation.

\begin{table}[]
\caption{Hearthstone Results}
\label{tab:hearthstone}
\begin{tabular}{l|cc|cc}
             & \multicolumn{2}{l|}{Maximum Overall}          & \multicolumn{2}{l}{All Solutions}  \\ 
    \toprule
Algorithm    & Fitness & Win Rate & Cells Filled & QD-Score \\
    \midrule

CMA-ES       & -2.471 & 44.8 \% & 17.02 \% & 16,024.8 \\
MAP-Elites   & -2.859 & 45.2 \% & 21.72 \% & 25,936.0 \\
ME (line)    & -0.252 & 50.5 \% & 22.30 \% & 28,132.7 \\
CMA-ME       & \bf{5.417} & \bf{65.9 \%} & \bf{29.17 \%} & \bf{63,295.6} \\    
  \bottomrule
\end{tabular}
\end{table}

\section{Discussion}
%TODO: mention figure 4b, potential of CMA-ES to be better Qd alg

% OLD - Both the toy domain and the Hearthstone domain share a property of the behavior space that challenges exploration with standard MAP-Elites: areas of the behavior space are hard to reach with random sampling. Particularly in the toy domain, discovering solutions in the tails of the behavior space formed by the Bates distribution described in Section~\ref{subsec:bates} would requires MAP-Elites to adapt its search distribution. While CMA-ES finds \textit{individual} solutions that outperform both QD algorithms, CMA-ES covers significantly less behavior space than any QD method.

Both the toy domain and the Hearthstone domain share a property of the behavior space that challenges exploration with standard MAP-Elites: some areas of the behavior space are hard to reach with random sampling even without the presence of deception. Particularly in the toy domain, discovering solutions in the tails of the behavior space formed by the Bates distribution described in Section~\ref{subsec:bates} requires MAP-Elites to adapt its search distribution. While CMA-ES finds \textit{individual} solutions that outperform both QD algorithms, in all of the chosen domains it covers significantly less area of the behavior space than any QD method.

% AMY - OLD Even in the Hearthstone domain, which unlike the toy domain is most likely a non-linear and non-separable mapping, it is challenging to explore the ``average number of turns'' dimension (horizontal axis in Fig.~\ref{fig:elite-map}). Because CMA-ME outperforms MAP-Elites in coverage (e.g., exploration), it is likely that an effect similar to the Bates distribution identified in the toy domain may be preventing MAP-Elites from discovering polices at the tails of behavior space. Interestingly, some behavioral characteristics like the ``average hand size'' do not share this property, illustrated by similar coverage across this dimension in both MAP-Elites and CMA-ME. Therefore, when selecting an algorithm, if maximal coverage is desired we recommend CMA-ME in part for its ability to efficiently explore a distorted behavior space. 

Because CMA-ME covers (e.g.,\ explores) more of the behavior space than MAP-Elites in the Hearthstone domain, even though the mapping is likely non-linear and non-separable, effects similar to those shown with the Bates distribution may be preventing \mbox{MAP-Elites} from discovering polices at the tails of behavior space like the ends of the ``average number of turns'' dimension (horizontal axis in Fig.~\ref{fig:elite-map}). However, some behavioral characteristics like the ``average hand size'' are fairly equally covered by the two algorithms. Therefore, when selecting an algorithm, if maximal coverage is desired we recommend CMA-ME in part for its ability to efficiently explore a distorted behavior space.  

%AMY: OLD -While it may be tempting to apply CMA-ES to any continuous optimization problem, results in Hearthstone suggest both \mbox{CMA-ME} and MAP-Elites better optimize for fitness (i.e.,\ win rate). It is possible that in this domain the objective function leads CMA-ES into a \emph{deceptive trap}, which has been observed in other domains~\cite{lehman:ec11}. One possible explanation is the best strategies discovered early are likely aggro, and that by discovering these strategies early in the search, CMA-ES consistently converges to the best aggro strategy instead of learning more complex control strategies. -- but tapers off as it converges -- Although this exploration is not sustained in the second half of the search where CMA-ES potentially encounters a \emph{deceptive trap} laid by the fitnes function in this domain~\cite{lehman:ec11}. 

While CMA-ES is often the algorithm of choice for solving problems in continuous optimization, Fig.~\ref{fig:rogue_win_rate} shows in Hearthstone that  \mbox{CMA-ME} and MAP-Elites better optimize for fitness (i.e.,\ win rate). Although interestingly in the first half of the search, Fig.~\ref{fig:rogue_qd_score} shows that CMA-ES has a higher QD-score than MAP-Elites, demonstrating its potential for exploration. While in the second half of search this exploration is not sustained as CMA-ES begins to converge, it is possible that in this domain the objective function leads CMA-ES into a \emph{deceptive trap}~\cite{lehman:ec11}. The best strategies discovered early are likely aggro, and that by discovering these strategies early in the search, CMA-ES consistently converges to the best aggro strategy instead of learning more complex control strategies.

While there are many possible emitter types, the CMA-ME algorithm proposes and explores three: optimizing, random direction, and improvement emitters. Because of their limited map interaction and ranking purely by fitness, optimizing emitters tend to explore a smaller area of behavior space. Random direction emitters alone can cover more of the behavior space, but solution quality is higher when searching with improvement emitters (Fig.~\ref{fig:elite-map}). Random emitters additionally require a direction in behavior space, which may be impossible for certain domains~\citep{khalifa:gecco19}. If diversity is more important than quality, we recommend random direction emitters, but alternatively recommend improvement emitters if defining a random direction vector is challenging or if the application prioritizes quality over diversity. We have only explored homogeneous emitter populations; we leave experimenting with heterogeneous populations of emitters as an exciting topic for future work. 

\section{Conclusions}

We presented a new algorithm called CMA-ME that combines the strengths of CMA-ES and MAP-Elites. Results from both a toy domain and a policy search in Hearthstone show that leveraging strengths from \mbox{CMA-ES} can improve both the coverage and quality of \mbox{MAP-Elites} solutions. Results from Hearthstone additionally show that when compared to standard CMA-ES, the diversity components from MAP-Elites can improve the quality of solutions in continuous search spaces. Overall, CMA-ME improves MAP-Elites by bringing modern optimization methods to quality-diversity problems for the first time. By reconceptualizing optimization problems as quality-diversity problems, results from CMA-ME suggest new opportunities for research and deployment, including any approach that learns policies encoded as neural networks; complementing the objective function with behavioral characteristics can yield not only a useful diversity of behavior, but also better performance on the original objective.

\begin{acks}
The authors would like to thank Heramb Nemlekar and Mark Nelson for their feedback on a preliminary version of this paper.

\end{acks}

\bibliographystyle{ACM-Reference-Format}
\bibliography{bibliography} 


\begin{thebibliography}{47}

%%% the Web address.

\ifx \showCODEN    \undefined \def \showCODEN     #1{\unskip}     \fi
\ifx \showDOI      \undefined \def \showDOI       #1{#1}\fi
\ifx \showISBNx    \undefined \def \showISBNx     #1{\unskip}     \fi
\ifx \showISBNxiii \undefined \def \showISBNxiii  #1{\unskip}     \fi
\ifx \showISSN     \undefined \def \showISSN      #1{\unskip}     \fi
\ifx \showLCCN     \undefined \def \showLCCN      #1{\unskip}     \fi
\ifx \shownote     \undefined \def \shownote      #1{#1}          \fi
\ifx \showarticletitle \undefined \def \showarticletitle #1{#1}   \fi
\ifx \showURL      \undefined \def \showURL       {\relax}        \fi
% invisible to TeX
\providecommand\bibfield[2]{#2}
\providecommand\bibinfo[2]{#2}
\providecommand\natexlab[1]{#1}
\providecommand\showeprint[2][]{arXiv:#2}

\bibitem[\protect\citeauthoryear{Alvarez, Dahlskog, Font, and Togelius}{Alvarez
  et~al\mbox{.}}{2019}]%
        {alvarez:cog19}
\bibfield{author}{\bibinfo{person}{Alberto Alvarez}, \bibinfo{person}{Steve
  Dahlskog}, \bibinfo{person}{Jose Font}, {and} \bibinfo{person}{Julian
  Togelius}.} \bibinfo{year}{2019}\natexlab{}.
\newblock \showarticletitle{Empowering Quality Diversity in Dungeon Design with
  Interactive Constrained MAP-Elites}.
\newblock \bibinfo{journal}{{\em IEEE Conference on Games (CoG)\/}}.
\newblock


\bibitem[\protect\citeauthoryear{Arulkumaran, Cully, and Togelius}{Arulkumaran
  et~al\mbox{.}}{2019}]%
        {arulkumaran:geccocomp19}
\bibfield{author}{\bibinfo{person}{Kai Arulkumaran}, \bibinfo{person}{Antoine
  Cully}, {and} \bibinfo{person}{Julian Togelius}.}
  \bibinfo{year}{2019}\natexlab{}.
\newblock \showarticletitle{AlphaStar: An Evolutionary Computation
  Perspective}. In \bibinfo{booktitle}{{\em GECCO '19: Proceedings of the
  Genetic and Evolutionary Computation Conference Companion}},
  \bibfield{editor}{\bibinfo{person}{Manuel L\'{o}pez-Ib\'{a}\~{n}ez}} (Ed.).
  \bibinfo{publisher}{ACM}, \bibinfo{address}{New York, NY, USA}.
\newblock
\showISBNx{978-1-4503-6748-6}


\bibitem[\protect\citeauthoryear{Bellemare, Naddaf, Veness, and
  Bowling}{Bellemare et~al\mbox{.}}{2013}]%
        {bellemare:ai13}
\bibfield{author}{\bibinfo{person}{Marc~G Bellemare}, \bibinfo{person}{Yavar
  Naddaf}, \bibinfo{person}{Joel Veness}, {and} \bibinfo{person}{Michael
  Bowling}.} \bibinfo{year}{2013}\natexlab{}.
\newblock \showarticletitle{The arcade learning environment: An evaluation
  platform for general agents}.
\newblock \bibinfo{journal}{{\em Journal of Artificial Intelligence
  Research\/}}  \bibinfo{volume}{47} (\bibinfo{year}{2013}),
  \bibinfo{pages}{253--279}.
\newblock


\bibitem[\protect\citeauthoryear{Bhatt, Lee, de~Mesentier~Silva, Watson,
  Togelius, and Hoover}{Bhatt et~al\mbox{.}}{2018}]%
        {bhatt:fdg18}
\bibfield{author}{\bibinfo{person}{Aditya Bhatt}, \bibinfo{person}{Scott Lee},
  \bibinfo{person}{Fernando de Mesentier~Silva}, \bibinfo{person}{Connor~W.
  Watson}, \bibinfo{person}{Julian Togelius}, {and} \bibinfo{person}{Amy~K.
  Hoover}.} \bibinfo{year}{2018}\natexlab{}.
\newblock \showarticletitle{Exploring the Hearthstone Deck Space}. In
  \bibinfo{booktitle}{{\em Proceedings of the 13th International Conference on
  the Foundations of Digital Games}}. \bibinfo{publisher}{ACM},
  \bibinfo{pages}{18}.
\newblock


\bibitem[\protect\citeauthoryear{Conti, Madhavan, Such, Lehman, Stanley, and
  Clune}{Conti et~al\mbox{.}}{2018}]%
        {conti:nips18}
\bibfield{author}{\bibinfo{person}{Edoardo Conti}, \bibinfo{person}{Vashisht
  Madhavan}, \bibinfo{person}{Felipe~Petroski Such}, \bibinfo{person}{Joel
  Lehman}, \bibinfo{person}{Kenneth~O. Stanley}, {and} \bibinfo{person}{Jeff
  Clune}.} \bibinfo{year}{2018}\natexlab{}.
\newblock \showarticletitle{Improving Exploration in Evolution Strategies for
  Deep Reinforcement Learning via a Population of Novelty-Seeking Agents}. In
  \bibinfo{booktitle}{{\em Proceedings of the 32Nd International Conference on
  Neural Information Processing Systems}} {\em (\bibinfo{series}{NIPS'18})}.
  \bibinfo{publisher}{Curran Associates Inc.}, \bibinfo{address}{USA},
  \bibinfo{pages}{5032--5043}.
\newblock
\showURL{%
\url{http://dl.acm.org/citation.cfm?id=3327345.3327410}}


\bibitem[\protect\citeauthoryear{Cuccu, Togelius, and Cudr{\'e}-Mauroux}{Cuccu
  et~al\mbox{.}}{2019}]%
        {cuccu:aamas19}
\bibfield{author}{\bibinfo{person}{Giuseppe Cuccu}, \bibinfo{person}{Julian
  Togelius}, {and} \bibinfo{person}{Philippe Cudr{\'e}-Mauroux}.}
  \bibinfo{year}{2019}\natexlab{}.
\newblock \showarticletitle{Playing atari with six neurons}. In
  \bibinfo{booktitle}{{\em Proceedings of the 18th International Conference on
  Autonomous Agents and MultiAgent Systems}}. International Foundation for
  Autonomous Agents and Multiagent Systems, \bibinfo{pages}{998--1006}.
\newblock


\bibitem[\protect\citeauthoryear{Cully}{Cully}{2019}]%
        {cully:gecco19}
\bibfield{author}{\bibinfo{person}{Antoine Cully}.}
  \bibinfo{year}{2019}\natexlab{}.
\newblock \showarticletitle{Autonomous Skill Discovery with Quality-Diversity
  and Unsupervised Descriptors}. In \bibinfo{booktitle}{{\em Proceedings of the
  Genetic and Evolutionary Computation Conference}} {\em
  (\bibinfo{series}{GECCO '19})}. \bibinfo{publisher}{ACM},
  \bibinfo{pages}{81--89}.
\newblock


\bibitem[\protect\citeauthoryear{Cully, Clune, Tarapore, and Mouret}{Cully
  et~al\mbox{.}}{2015}]%
        {cully:nature15}
\bibfield{author}{\bibinfo{person}{Antoine Cully}, \bibinfo{person}{Jeff
  Clune}, \bibinfo{person}{Danesh Tarapore}, {and}
  \bibinfo{person}{Jean-Baptiste Mouret}.} \bibinfo{year}{2015}\natexlab{}.
\newblock \showarticletitle{Robots that can adapt like animals}.
\newblock \bibinfo{journal}{{\em Nature\/}} \bibinfo{volume}{521},
  \bibinfo{number}{7553} (\bibinfo{year}{2015}), \bibinfo{pages}{503}.
\newblock


\bibitem[\protect\citeauthoryear{Cully and Mouret}{Cully and Mouret}{2013}]%
        {cully:gecco13}
\bibfield{author}{\bibinfo{person}{Antoine Cully} {and}
  \bibinfo{person}{Jean-Baptiste Mouret}.} \bibinfo{year}{2013}\natexlab{}.
\newblock \showarticletitle{Behavioral Repertoire Learning in Robotics}. In
  \bibinfo{booktitle}{{\em Proceedings of the 15th Annual Conference on Genetic
  and Evolutionary Computation (GECCO ‘13)}}. \bibinfo{publisher}{ACM},
  \bibinfo{pages}{175–--182}.
\newblock


\bibitem[\protect\citeauthoryear{Cully and Mouret}{Cully and Mouret}{2016}]%
        {cully:ec16}
\bibfield{author}{\bibinfo{person}{Antoine Cully} {and}
  \bibinfo{person}{Jean-Baptiste Mouret}.} \bibinfo{year}{2016}\natexlab{}.
\newblock \showarticletitle{Evolving a Behavioral Repertoire for a Walking
  Robot}.
\newblock \bibinfo{journal}{{\em Evolutionary Computation\/}}
  \bibinfo{volume}{24} (\bibinfo{year}{2016}), \bibinfo{pages}{59--88}.
\newblock
Issue 1.


\bibitem[\protect\citeauthoryear{de~Mesentier~Silva, Canaan, Lee, Fontaine,
  Togelius, and Hoover}{de~Mesentier~Silva et~al\mbox{.}}{2019}]%
        {de2019evolving}
\bibfield{author}{\bibinfo{person}{Fernando de Mesentier~Silva},
  \bibinfo{person}{Rodrigo Canaan}, \bibinfo{person}{Scott Lee},
  \bibinfo{person}{Matthew~C Fontaine}, \bibinfo{person}{Julian Togelius},
  {and} \bibinfo{person}{Amy~K Hoover}.} \bibinfo{year}{2019}\natexlab{}.
\newblock \showarticletitle{Evolving the hearthstone meta}. In
  \bibinfo{booktitle}{{\em 2019 IEEE Conference on Games (CoG)}}. IEEE,
  \bibinfo{pages}{1--8}.
\newblock


\bibitem[\protect\citeauthoryear{Decoster, Choe, et~al\mbox{.}}{Decoster
  et~al\mbox{.}}{2019}]%
        {sabberstone:github19}
\bibfield{author}{\bibinfo{person}{Cedric Decoster}, \bibinfo{person}{Jean
  Seong~Bjorn Choe}, {et~al\mbox{.}}} \bibinfo{year}{2019}\natexlab{}.
\newblock \bibinfo{booktitle}{{\em Sabberstone}}.
\newblock
\showURL{%
\url{https://github.com/HearthSim/SabberStone}}
\newblock
\shownote{Accessed: 2019-11-01.}


\bibitem[\protect\citeauthoryear{{Eiben}, {Hinterding}, and
  {Michalewicz}}{{Eiben} et~al\mbox{.}}{1999}]%
        {eiben:ec99}
\bibfield{author}{\bibinfo{person}{A.~E. {Eiben}}, \bibinfo{person}{R.
  {Hinterding}}, {and} \bibinfo{person}{Z. {Michalewicz}}.}
  \bibinfo{year}{1999}\natexlab{}.
\newblock \showarticletitle{Parameter control in evolutionary algorithms}.
\newblock \bibinfo{journal}{{\em IEEE Transactions on Evolutionary
  Computation\/}} \bibinfo{volume}{3}, \bibinfo{number}{2}
  (\bibinfo{date}{July} \bibinfo{year}{1999}), \bibinfo{pages}{124--141}.
\newblock
\showDOI{%
\url{https://doi.org/10.1109/4235.771166}}


\bibitem[\protect\citeauthoryear{Entertainment}{Entertainment}{[n. d.]}]%
        {hearthstone:web19}
\bibfield{author}{\bibinfo{person}{Blizzard Entertainment}.} \bibinfo{year}{[n.
  d.]}\natexlab{}.
\newblock \bibinfo{title}{Hearthstone}.
\newblock \bibinfo{howpublished}{\url{https://playhearthstone.com/en-us/}}.
  (\bibinfo{year}{[n. d.]}).
\newblock
\newblock
\shownote{Accessed: 2019-11-01.}


\bibitem[\protect\citeauthoryear{Fioravanzo and Iacca}{Fioravanzo and
  Iacca}{2019}]%
        {fioravanzo:arxiv19}
\bibfield{author}{\bibinfo{person}{Stefano Fioravanzo} {and}
  \bibinfo{person}{Giovanni Iacca}.} \bibinfo{year}{2019}\natexlab{}.
\newblock \bibinfo{title}{Evaluating MAP-Elites on Constrained Optimization
  Problems}.
\newblock   (\bibinfo{year}{2019}).
\newblock
\showeprint{1902.00703}


\bibitem[\protect\citeauthoryear{Fontaine}{Fontaine}{2019a}]%
        {evostone:github19}
\bibfield{author}{\bibinfo{person}{Matthew Fontaine}.}
  \bibinfo{year}{2019}\natexlab{a}.
\newblock \bibinfo{booktitle}{{\em EvoStone}}.
\newblock
\showURL{%
\url{https://github.com/tehqin/EvoStone}}
\newblock
\shownote{Accessed: 2019-12-01.}


\bibitem[\protect\citeauthoryear{Fontaine}{Fontaine}{2019b}]%
        {qd-benchmark:github19}
\bibfield{author}{\bibinfo{person}{Matthew Fontaine}.}
  \bibinfo{year}{2019}\natexlab{b}.
\newblock \bibinfo{booktitle}{{\em QualDivBenchmark}}.
\newblock
\showURL{%
\url{https://github.com/tehqin/QualDivBenchmark}}
\newblock
\shownote{Accessed: 2019-12-01.}


\bibitem[\protect\citeauthoryear{Fontaine, Lee, Soros, de~Mesentier~Silva,
  Togelius, and Hoover}{Fontaine et~al\mbox{.}}{2019}]%
        {fontaine:gecco19}
\bibfield{author}{\bibinfo{person}{Matthew~C. Fontaine}, \bibinfo{person}{Scott
  Lee}, \bibinfo{person}{L.~B. Soros}, \bibinfo{person}{Fernando de
  Mesentier~Silva}, \bibinfo{person}{Julian Togelius}, {and}
  \bibinfo{person}{Amy~K. Hoover}.} \bibinfo{year}{2019}\natexlab{}.
\newblock \showarticletitle{Mapping Hearthstone Deck Spaces through MAP-Elites
  with Sliding Boundaries}. In \bibinfo{booktitle}{{\em Proceedings of the
  Genetic and Evolutionary Computation Conference}} {\em
  (\bibinfo{series}{GECCO '19})}. \bibinfo{publisher}{ACM},
  \bibinfo{address}{New York, NY, USA}, \bibinfo{pages}{161--169}.
\newblock
\showISBNx{978-1-4503-6111-8}
\showDOI{%
\url{https://doi.org/10.1145/3321707.3321794}}


\bibitem[\protect\citeauthoryear{Gaier, Asteroth, and Mouret}{Gaier
  et~al\mbox{.}}{2019}]%
        {gaier:gecco19}
\bibfield{author}{\bibinfo{person}{Adam Gaier}, \bibinfo{person}{Alexander
  Asteroth}, {and} \bibinfo{person}{Jean-Baptiste Mouret}.}
  \bibinfo{year}{2019}\natexlab{}.
\newblock \showarticletitle{Are Quality Diversity Algorithms Better at
  Generating Stepping Stones than Objective-Based Search?}. In
  \bibinfo{booktitle}{{\em Proceedings of the Genetic and Evolutionary
  Computation Conference Companion}}. ACM, \bibinfo{pages}{115--116}.
\newblock


\bibitem[\protect\citeauthoryear{Garc{\'\i}a-S{\'a}nchez, Tonda,
  Fern{\'a}ndez-Leiva, and Cotta}{Garc{\'\i}a-S{\'a}nchez
  et~al\mbox{.}}{2019}]%
        {garcia:kbs19}
\bibfield{author}{\bibinfo{person}{Pablo Garc{\'\i}a-S{\'a}nchez},
  \bibinfo{person}{Alberto Tonda}, \bibinfo{person}{Antonio~J
  Fern{\'a}ndez-Leiva}, {and} \bibinfo{person}{Carlos Cotta}.}
  \bibinfo{year}{2019}\natexlab{}.
\newblock \showarticletitle{Optimizing Hearthstone agents using an evolutionary
  algorithm}.
\newblock \bibinfo{journal}{{\em Knowledge-Based Systems\/}}
  (\bibinfo{year}{2019}), \bibinfo{pages}{105032}.
\newblock


\bibitem[\protect\citeauthoryear{Glasmachers, Schaul, Yi, Wierstra, and
  Schmidhuber}{Glasmachers et~al\mbox{.}}{2010}]%
        {glasmachers:gecco10}
\bibfield{author}{\bibinfo{person}{Tobias Glasmachers}, \bibinfo{person}{Tom
  Schaul}, \bibinfo{person}{Sun Yi}, \bibinfo{person}{Daan Wierstra}, {and}
  \bibinfo{person}{J{\"u}rgen Schmidhuber}.} \bibinfo{year}{2010}\natexlab{}.
\newblock \showarticletitle{Exponential natural evolution strategies}. In
  \bibinfo{booktitle}{{\em Proceedings of the 12th annual conference on Genetic
  and evolutionary computation}}. ACM, \bibinfo{pages}{393--400}.
\newblock


\bibitem[\protect\citeauthoryear{Gravina, Khalifa, Liapis, Togelius, and
  Yannakakis}{Gravina et~al\mbox{.}}{2019}]%
        {gravina:cog19}
\bibfield{author}{\bibinfo{person}{Daniele Gravina}, \bibinfo{person}{Ahmed
  Khalifa}, \bibinfo{person}{Antonios Liapis}, \bibinfo{person}{Julian
  Togelius}, {and} \bibinfo{person}{Georgios Yannakakis}.}
  \bibinfo{year}{2019}\natexlab{}.
\newblock \showarticletitle{Procedural Content Generation through Quality
  Diversity}.
\newblock \bibinfo{journal}{{\em IEEE Conference on Games (CoG)\/}},
  \bibinfo{pages}{1--8}.
\newblock
\showDOI{%
\url{https://doi.org/10.1109/CIG.2019.8848053}}


\bibitem[\protect\citeauthoryear{Hansen}{Hansen}{2016}]%
        {hansen:cma16}
\bibfield{author}{\bibinfo{person}{Nikolaus Hansen}.}
  \bibinfo{year}{2016}\natexlab{}.
\newblock \showarticletitle{The CMA evolution strategy: A tutorial}.
\newblock \bibinfo{journal}{{\em arXiv preprint arXiv:1604.00772\/}}
  (\bibinfo{year}{2016}).
\newblock


\bibitem[\protect\citeauthoryear{Hansen, Auger, Mersmann, Tusar, and
  Brockhoff}{Hansen et~al\mbox{.}}{2016}]%
        {hansen:arxiv16}
\bibfield{author}{\bibinfo{person}{N. Hansen}, \bibinfo{person}{A. Auger},
  \bibinfo{person}{O. Mersmann}, \bibinfo{person}{T. Tusar}, {and}
  \bibinfo{person}{D. Brockhoff}.} \bibinfo{year}{2016}\natexlab{}.
\newblock \showarticletitle{COCO: A platform for comparing continuous
  optimizers in a black-box setting}.
\newblock  (\bibinfo{year}{2016}).
\newblock


\bibitem[\protect\citeauthoryear{Hansen, Auger, Ros, Finck, and Pošík}{Hansen
  et~al\mbox{.}}{2010}]%
        {hansen:gecco10}
\bibfield{author}{\bibinfo{person}{Nikolaus Hansen}, \bibinfo{person}{Anne
  Auger}, \bibinfo{person}{Raymond Ros}, \bibinfo{person}{Steffen Finck}, {and}
  \bibinfo{person}{Petr Pošík}.} \bibinfo{year}{2010}\natexlab{}.
\newblock \showarticletitle{Comparing Results of 31 Algorithms from the
  Black-Box Optimization Benchmarking BBOB-2009}.
\newblock \bibinfo{journal}{{\em Proceedings of the 12th Annual Genetic and
  Evolutionary Computation Conference, GECCO '10 - Companion Publication\/}},
  \bibinfo{pages}{1689--1696}.
\newblock
\showDOI{%
\url{https://doi.org/10.1145/1830761.1830790}}


\bibitem[\protect\citeauthoryear{Hansen and Kern}{Hansen and Kern}{2004}]%
        {hansen2004evaluating}
\bibfield{author}{\bibinfo{person}{Nikolaus Hansen} {and}
  \bibinfo{person}{Stefan Kern}.} \bibinfo{year}{2004}\natexlab{}.
\newblock \showarticletitle{Evaluating the CMA evolution strategy on multimodal
  test functions}. In \bibinfo{booktitle}{{\em International Conference on
  Parallel Problem Solving from Nature}}. Springer, \bibinfo{pages}{282--291}.
\newblock


\bibitem[\protect\citeauthoryear{Hansen and Ostermeier}{Hansen and
  Ostermeier}{2001}]%
        {hansen:ec01}
\bibfield{author}{\bibinfo{person}{Nikolaus Hansen} {and}
  \bibinfo{person}{Andreas Ostermeier}.} \bibinfo{year}{2001}\natexlab{}.
\newblock \showarticletitle{Completely derandomized self-adaptation in
  evolution strategies}.
\newblock \bibinfo{journal}{{\em Evolutionary computation\/}}
  \bibinfo{volume}{9}, \bibinfo{number}{2} (\bibinfo{year}{2001}),
  \bibinfo{pages}{159--195}.
\newblock


\bibitem[\protect\citeauthoryear{Hoover, Togelius, Lee, and
  de~Mesentier~Silva}{Hoover et~al\mbox{.}}{2019}]%
        {hoover:AI19}
\bibfield{author}{\bibinfo{person}{Amy~K Hoover}, \bibinfo{person}{Julian
  Togelius}, \bibinfo{person}{Scott Lee}, {and} \bibinfo{person}{Fernando de
  Mesentier~Silva}.} \bibinfo{year}{2019}\natexlab{}.
\newblock \showarticletitle{The Many AI Challenges of Hearthstone}.
\newblock \bibinfo{journal}{{\em KI-K{\"u}nstliche Intelligenz\/}}
  (\bibinfo{year}{2019}), \bibinfo{pages}{1--11}.
\newblock


\bibitem[\protect\citeauthoryear{HSReplay}{HSReplay}{[n. d.]}]%
        {hsreplay:web19}
\bibfield{author}{\bibinfo{person}{HSReplay}.} \bibinfo{year}{[n.
  d.]}\natexlab{}.
\newblock \bibinfo{title}{HSReplay}.
\newblock \bibinfo{howpublished}{\url{https://hsreplay.net/}}.
  (\bibinfo{year}{[n. d.]}).
\newblock
\newblock
\shownote{Accessed: 2019-11-01.}


\bibitem[\protect\citeauthoryear{Khalifa, Green, Barros, and Togelius}{Khalifa
  et~al\mbox{.}}{2019}]%
        {khalifa:gecco19}
\bibfield{author}{\bibinfo{person}{Ahmed Khalifa},
  \bibinfo{person}{Michael~Cerny Green}, \bibinfo{person}{Gabriella Barros},
  {and} \bibinfo{person}{Julian Togelius}.} \bibinfo{year}{2019}\natexlab{}.
\newblock \showarticletitle{Intentional computational level design}. In
  \bibinfo{booktitle}{{\em Proceedings of The Genetic and Evolutionary
  Computation Conference}}. \bibinfo{pages}{796--803}.
\newblock


\bibitem[\protect\citeauthoryear{Khalifa, Lee, Nealen, and Togelius}{Khalifa
  et~al\mbox{.}}{2018}]%
        {khalifa:gecco18}
\bibfield{author}{\bibinfo{person}{Ahmed Khalifa}, \bibinfo{person}{Scott Lee},
  \bibinfo{person}{Andy Nealen}, {and} \bibinfo{person}{Julian Togelius}.}
  \bibinfo{year}{2018}\natexlab{}.
\newblock \showarticletitle{Talakat: Bullet Hell Generation Through Constrained
  Map-elites}. In \bibinfo{booktitle}{{\em Proceedings of the Genetic and
  Evolutionary Computation Conference}} {\em (\bibinfo{series}{GECCO '18})}.
  \bibinfo{publisher}{ACM}, \bibinfo{address}{New York, NY, USA},
  \bibinfo{pages}{1047--1054}.
\newblock
\showISBNx{978-1-4503-5618-3}
\showDOI{%
\url{https://doi.org/10.1145/3205455.3205470}}


\bibitem[\protect\citeauthoryear{Lehman and Stanley}{Lehman and
  Stanley}{2008}]%
        {lehman:alife08}
\bibfield{author}{\bibinfo{person}{Joel Lehman} {and}
  \bibinfo{person}{Kenneth~O. Stanley}.} \bibinfo{year}{2008}\natexlab{}.
\newblock \showarticletitle{Exploiting Open-Endedness to Solve Problems through
  the Search for Novelty}. In \bibinfo{booktitle}{{\em Proceedings of the
  Eleventh International Conference on Artificial Life (Alife XI)}}.
  \bibinfo{pages}{329--336}.
\newblock


\bibitem[\protect\citeauthoryear{Lehman and Stanley}{Lehman and
  Stanley}{2011a}]%
        {lehman:ec11}
\bibfield{author}{\bibinfo{person}{Joel Lehman} {and}
  \bibinfo{person}{Kenneth~O. Stanley}.} \bibinfo{year}{2011}\natexlab{a}.
\newblock \showarticletitle{Abandoning Objectives: Evolution through the Search
  for Novelty Alone}.
\newblock \bibinfo{journal}{{\em Evolutionary Computation\/}}
  (\bibinfo{year}{2011}).
\newblock


\bibitem[\protect\citeauthoryear{Lehman and Stanley}{Lehman and
  Stanley}{2011b}]%
        {lehman:gecco11}
\bibfield{author}{\bibinfo{person}{Joel Lehman} {and}
  \bibinfo{person}{Kenneth~O. Stanley}.} \bibinfo{year}{2011}\natexlab{b}.
\newblock \showarticletitle{Evolving a Diversity of Virtual Creatures through
  Novelty Search and Local Competition}. In \bibinfo{booktitle}{{\em
  Proceedings of the 13th Annual Conference on Genetic and Evolutionary
  Computation (GECCO ‘11)}}. \bibinfo{publisher}{ACM},
  \bibinfo{pages}{211–--218}.
\newblock


\bibitem[\protect\citeauthoryear{Liang}{Liang}{[n. d.]}]%
        {roguedeck:web19}
\bibfield{author}{\bibinfo{person}{Fei Liang}.} \bibinfo{year}{[n.
  d.]}\natexlab{}.
\newblock \bibinfo{title}{Tempo Rogue}.
\newblock
  \bibinfo{howpublished}{\url{https://www.hearthstonetopdecks.com/decks/tempo-rogue-rise-of-shadows-1-legend-etc/}}.
    (\bibinfo{year}{[n. d.]}).
\newblock
\newblock
\shownote{Accessed: 2019-11-01.}


\bibitem[\protect\citeauthoryear{Ng, Harada, and Russell}{Ng
  et~al\mbox{.}}{1999}]%
        {ng:icml99}
\bibfield{author}{\bibinfo{person}{Andrew~Y. Ng}, \bibinfo{person}{Daishi
  Harada}, {and} \bibinfo{person}{Stuart~J. Russell}.}
  \bibinfo{year}{1999}\natexlab{}.
\newblock \showarticletitle{Policy Invariance Under Reward Transformations:
  Theory and Application to Reward Shaping}. In \bibinfo{booktitle}{{\em
  Proceedings of the Sixteenth International Conference on Machine Learning}}
  {\em (\bibinfo{series}{ICML '99})}. \bibinfo{publisher}{Morgan Kaufmann
  Publishers Inc.}, \bibinfo{address}{San Francisco, CA, USA},
  \bibinfo{pages}{278--287}.
\newblock
\showISBNx{1-55860-612-2}
\showURL{%
\url{http://dl.acm.org/citation.cfm?id=645528.657613}}


\bibitem[\protect\citeauthoryear{Nordmoen, Ellefsen, and Glette}{Nordmoen
  et~al\mbox{.}}{2018a}]%
        {nordmoen:ec18}
\bibfield{author}{\bibinfo{person}{Jørgen Nordmoen}, \bibinfo{person}{Kai~Olav
  Ellefsen}, {and} \bibinfo{person}{Kyrre Glette}.}
  \bibinfo{year}{2018}\natexlab{a}.
\newblock \bibinfo{booktitle}{{\em Combining MAP-Elites and Incremental
  Evolution to Generate Gaits for a Mammalian Quadruped Robot}}.
\newblock \bibinfo{pages}{719--733}.
\newblock
\showISBNx{978-3-319-77537-1}
\showDOI{%
\url{https://doi.org/10.1007/978-3-319-77538-8_48}}


\bibitem[\protect\citeauthoryear{Nordmoen, Samuelsen, Ellefsen, and
  Glette}{Nordmoen et~al\mbox{.}}{2018b}]%
        {nordmoen:alife18}
\bibfield{author}{\bibinfo{person}{J{\o}rgen Nordmoen}, \bibinfo{person}{Eivind
  Samuelsen}, \bibinfo{person}{Kai~Olav Ellefsen}, {and} \bibinfo{person}{Kyrre
  Glette}.} \bibinfo{year}{2018}\natexlab{b}.
\newblock \showarticletitle{Dynamic Mutation in MAP-Elites for Robotic
  Repertoire Generation}. In \bibinfo{booktitle}{{\em Artificial Life
  Conference Proceedings}}. \bibinfo{publisher}{MIT Press},
  \bibinfo{pages}{598--605}.
\newblock
\showDOI{%
\url{https://doi.org/10.1162/isal\_a\_00110}}


\bibitem[\protect\citeauthoryear{Preuss}{Preuss}{2012}]%
        {preuss:ecec12}
\bibfield{author}{\bibinfo{person}{Mike Preuss}.}
  \bibinfo{year}{2012}\natexlab{}.
\newblock \showarticletitle{Improved Topological Niching for Real-valued Global
  Optimization}. In \bibinfo{booktitle}{{\em Proceedings of the 2012T European
  Conference on Applications of Evolutionary Computation}} {\em
  (\bibinfo{series}{EvoApplications'12})}.
  \bibinfo{publisher}{Springer-Verlag}, \bibinfo{address}{Berlin, Heidelberg},
  \bibinfo{pages}{386--395}.
\newblock
\showISBNx{978-3-642-29177-7}
\showDOI{%
\url{https://doi.org/10.1007/978-3-642-29178-4_39}}


\bibitem[\protect\citeauthoryear{Pugh, Soros, and Stanley}{Pugh
  et~al\mbox{.}}{2016}]%
        {pugh:frontiers16}
\bibfield{author}{\bibinfo{person}{Justin~K. Pugh}, \bibinfo{person}{Lisa~B.
  Soros}, {and} \bibinfo{person}{Kenneth~O. Stanley}.}
  \bibinfo{year}{2016}\natexlab{}.
\newblock \showarticletitle{Quality Diversity: A New Frontier for Evolutionary
  Computation}.
\newblock \bibinfo{journal}{{\em Frontiers in Robotics and AI\/}}
  \bibinfo{volume}{3} (\bibinfo{year}{2016}), \bibinfo{pages}{40}.
\newblock
\showISSN{2296-9144}
\showDOI{%
\url{https://doi.org/10.3389/frobt.2016.00040}}


\bibitem[\protect\citeauthoryear{Pugh, Soros, Szerlip, and Stanley}{Pugh
  et~al\mbox{.}}{2015}]%
        {pugh:gecco15}
\bibfield{author}{\bibinfo{person}{Justin~K. Pugh}, \bibinfo{person}{L.~B.
  Soros}, \bibinfo{person}{Paul~A. Szerlip}, {and} \bibinfo{person}{Kenneth~O.
  Stanley}.} \bibinfo{year}{2015}\natexlab{}.
\newblock \showarticletitle{Confronting the Challenge of Quality Diversity}. In
  \bibinfo{booktitle}{{\em Proceedings of the 2015 Annual Conference on Genetic
  and Evolutionary Computation}} {\em (\bibinfo{series}{GECCO '15})}.
  \bibinfo{publisher}{ACM}, \bibinfo{address}{New York, NY, USA},
  \bibinfo{pages}{967--974}.
\newblock
\showISBNx{978-1-4503-3472-3}
\showDOI{%
\url{https://doi.org/10.1145/2739480.2754664}}


\bibitem[\protect\citeauthoryear{Shir and B\"{a}ck}{Shir and B\"{a}ck}{2006}]%
        {shir:icpps06}
\bibfield{author}{\bibinfo{person}{Ofer~M. Shir} {and} \bibinfo{person}{Thomas
  B\"{a}ck}.} \bibinfo{year}{2006}\natexlab{}.
\newblock \showarticletitle{Niche Radius Adaptation in the CMA-ES Niching
  Algorithm}. In \bibinfo{booktitle}{{\em Proceedings of the 9th International
  Conference on Parallel Problem Solving from Nature}} {\em
  (\bibinfo{series}{PPSN'06})}. \bibinfo{publisher}{Springer-Verlag},
  \bibinfo{address}{Berlin, Heidelberg}, \bibinfo{pages}{142--151}.
\newblock
\showISBNx{3-540-38990-3, 978-3-540-38990-3}
\showDOI{%
\url{https://doi.org/10.1007/11844297_15}}


\bibitem[\protect\citeauthoryear{Smith, Tokarchuk, and Wiggins}{Smith
  et~al\mbox{.}}{2016}]%
        {smith:ppsn16}
\bibfield{author}{\bibinfo{person}{Davy Smith}, \bibinfo{person}{Laurissa
  Tokarchuk}, {and} \bibinfo{person}{Geraint Wiggins}.}
  \bibinfo{year}{2016}\natexlab{}.
\newblock \showarticletitle{Rapid phenotypic landscape exploration through
  hierarchical spatial partitioning}. In \bibinfo{booktitle}{{\em International
  conference on parallel problem solving from nature}}. Springer,
  \bibinfo{pages}{911--920}.
\newblock


\bibitem[\protect\citeauthoryear{{\'S}wiechowski, Tajmajer, and
  Janusz}{{\'S}wiechowski et~al\mbox{.}}{2018}]%
        {swiechowski:cig18}
\bibfield{author}{\bibinfo{person}{Maciej {\'S}wiechowski},
  \bibinfo{person}{Tomasz Tajmajer}, {and} \bibinfo{person}{Andrzej Janusz}.}
  \bibinfo{year}{2018}\natexlab{}.
\newblock \showarticletitle{Improving Hearthstone AI by Combining MCTS and
  Supervised Learning Algorithms}. In \bibinfo{booktitle}{{\em 2018 IEEE
  Conference on Computational Intelligence and Games (CIG)}}.
  \bibinfo{publisher}{IEEE}, \bibinfo{pages}{1--8}.
\newblock


\bibitem[\protect\citeauthoryear{Vassiliades, Chatzilygeroudis, and
  Mouret}{Vassiliades et~al\mbox{.}}{2018}]%
        {vassiliades:ec18}
\bibfield{author}{\bibinfo{person}{Vassilis Vassiliades},
  \bibinfo{person}{Konstantinos Chatzilygeroudis}, {and}
  \bibinfo{person}{Jean-Baptiste Mouret}.} \bibinfo{year}{2018}\natexlab{}.
\newblock \showarticletitle{Using Centroidal Voronoi Tessellations to Scale Up
  the Multidimensional Archive of Phenotypic Elites Algorithm}.
\newblock \bibinfo{journal}{{\em IEEE Transactions on Evolutionary
  Computation\/}} \bibinfo{volume}{22}, \bibinfo{number}{4}
  (\bibinfo{year}{2018}), \bibinfo{pages}{623--630}.
\newblock


\bibitem[\protect\citeauthoryear{Vassiliades and Mouret}{Vassiliades and
  Mouret}{2018}]%
        {vassiliades:gecco18}
\bibfield{author}{\bibinfo{person}{Vassiiis Vassiliades} {and}
  \bibinfo{person}{Jean-Baptiste Mouret}.} \bibinfo{year}{2018}\natexlab{}.
\newblock \showarticletitle{Discovering the elite hypervolume by leveraging
  interspecies correlation}. In \bibinfo{booktitle}{{\em Proceedings of the
  Genetic and Evolutionary Computation Conference}}. \bibinfo{pages}{149--156}.
\newblock


\bibitem[\protect\citeauthoryear{Vinyals, Babuschkin, Czarnecki, Mathieu,
  Dudzik, Chung, Choi, Powell, Ewalds, Georgiev, et~al\mbox{.}}{Vinyals
  et~al\mbox{.}}{2019}]%
        {yinyals:nature19}
\bibfield{author}{\bibinfo{person}{Oriol Vinyals}, \bibinfo{person}{Igor
  Babuschkin}, \bibinfo{person}{Wojciech~M. Czarnecki},
  \bibinfo{person}{Micha{\"e}l Mathieu}, \bibinfo{person}{Andrew Dudzik},
  \bibinfo{person}{Junyoung Chung}, \bibinfo{person}{David~H. Choi},
  \bibinfo{person}{Richard Powell}, \bibinfo{person}{Timo Ewalds},
  \bibinfo{person}{Petko Georgiev}, {et~al\mbox{.}}}
  \bibinfo{year}{2019}\natexlab{}.
\newblock \showarticletitle{Grandmaster level in StarCraft II using multi-agent
  reinforcement learning}.
\newblock \bibinfo{journal}{{\em Nature\/}} \bibinfo{volume}{575},
  \bibinfo{number}{11} (\bibinfo{year}{2019}), \bibinfo{pages}{350--354}.
\newblock


\end{thebibliography}

\end{document}